\documentclass{article}
\usepackage[T1]{fontenc}
\usepackage{iclr2027_conference,times}
\usepackage{hyperref}
\usepackage{url}
\usepackage{graphicx}
\usepackage{amsmath,amssymb,amsthm}
\usepackage{flafter}
\usepackage{tabularx}
\usepackage{xcolor}
\usepackage{colortbl}
\usepackage{booktabs}
\usepackage{caption}
\definecolor{ours}{HTML}{D55E00}
\definecolor{oursbg}{HTML}{FBEFE6}
\definecolor{link}{HTML}{1F4E8C}
\definecolor{ink}{HTML}{3A3A3A}
\definecolor{pours}{HTML}{F8E5D6}
\definecolor{pblue}{HTML}{D6E8F3}
\definecolor{pgreen}{HTML}{D6EFE9}
\definecolor{ppurple}{HTML}{F5E4ED}
\definecolor{pgray}{HTML}{EDEDED}
\definecolor{mhybrid}{HTML}{7C3700}
\definecolor{mranddepth}{HTML}{0072B2}
\definecolor{mmeandepth}{HTML}{009E73}
\definecolor{mwidth}{HTML}{CC79A7}
\definecolor{mstandard}{HTML}{8C8C8C}
\definecolor{graytext}{HTML}{222222}
\definecolor{graydark}{HTML}{4D4D4D}
\definecolor{graymid}{HTML}{8C8C8C}
\definecolor{graylight}{HTML}{BFBFBF}
\definecolor{grayfaint}{HTML}{EDEDED}
\newcommand{\std}[1]{{\scriptsize\,$\pm$#1}}
\usepackage{tikz}
\usetikzlibrary{arrows.meta,positioning,fit,backgrounds,calc}
\tikzset{
  fs box/.style={rounded corners=3pt, draw=none, fill=pblue, text=ink, align=center,
                 minimum height=7mm, inner xsep=5pt, inner ysep=3pt, font=\small},
  fs ours/.style={fs box, fill=pours},
  fs base/.style={fs box, fill=pgray},
  fs group/.style={rounded corners=6pt, fill=grayfaint!60, draw=none},
  fs arrow/.style={-{Stealth[length=2.2mm,width=1.6mm]}, line width=0.8pt, draw=ink},
  fs dashed/.style={fs arrow, dashed, draw=graymid},
  fs note/.style={font=\footnotesize\itshape, text=graydark, align=center},
  fs good/.style={font=\footnotesize\itshape, text=ours},
}
\usepackage[skins,breakable]{tcolorbox}
\newtcolorbox{takeaway}[1][Takeaways]{enhanced, breakable, colback=link!4, colframe=link!45,
  boxrule=0.4pt, arc=1.5pt, left=5pt, right=5pt, top=3pt, bottom=3pt, fontupper=\small,
  before upper={{\bfseries\color{link!80!black}#1}\par\smallskip}}
\newcommand{\ourmethod}{\textsc{OurMethod}}  % 手法名の表記を 1 か所で管理 (名前を書き換える)
\graphicspath{{figures/}}
\hypersetup{colorlinks=true, linkcolor=link, citecolor=link, urlcolor=link}

\renewcommand{\ourmethod}{BoD}
\newcommand{\cL}{\mathcal{L}}
\newcommand{\Dtr}{D_{\mathrm{train}}}
\newcommand{\Dte}{D_{\mathrm{test}}}
\newcommand{\Enc}{\mathrm{Enc}}
\newcommand{\Dec}{\mathrm{Dec}}
\newcommand{\1}{\mathbf{1}}

\newtheorem{lemma}{Lemma}
\newtheoremstyle{boxedproposition}{0pt}{0pt}{\itshape}{}{\bfseries}{.}{.5em}{}
\theoremstyle{boxedproposition}
\newtheorem{proposition}{Proposition}
\theoremstyle{plain}
\newtcolorbox{theorybox}{enhanced, colback=black!2, colframe=graylight,
  boxrule=0.4pt, arc=1pt, left=6pt, right=6pt, top=2pt, bottom=2pt,
  before skip=4pt, after skip=4pt}

\title{Best-of-Depth: Learning Exploratory\\Dynamics in Looped Transformers}
\author{%
\makebox[\linewidth][c]{%
\begin{tabular}{@{}c@{}}
Shota Takashiro ~~ Takeshi Kojima ~~ Masanori Koyama\\[0.3em]
 Kohei Hayashi ~~ Yusuke Iwasawa
\\[0.5em]
\normalfont{The University of Tokyo}\\
\normalfont{takashiro@weblab.t.u-tokyo.ac.jp}
\end{tabular}%
}
}

\iclrfinalcopy

\begin{document}
\maketitle
\lhead{Under review as a conference paper at ICLR 2027}

% =============================================================================
\begin{abstract}
Looped Transformers extend inference-time computation through repeated latent-state updates,
but final-depth training does not directly encourage useful intermediate alternatives. We introduce
Best-of-Depth (\ourmethod{}), which trains a recurrent trajectory as a candidate set by minimizing
prediction loss across depths. Different valid targets can be matched at different depths without
architectural changes or depth-specific labels. An exact-match analysis connects this objective
to solution coverage. On N-Queens completion and graph coloring, BoD improves coverage
over final-depth and intermediate-supervision objectives. N-Queens coverage reaches 91.03\%,
versus 45.75\% for final-depth training. Further experiments show discovery
beyond the training depth and recovery of solutions withheld from teacher pools on familiar inputs.
Fine-tuning a pretrained looped language model with BoD yields higher pooled code-generation
success than final-depth training when programs are generated separately at evaluated fixed depths.
These results support training recurrent
computation for solution discovery along its trajectory.
\end{abstract}

% =============================================================================
\section{Introduction}
\label{sec:intro}

Additional inference-time computation can improve a model's ability to solve problems
\citep{snell2025scaling,brown2024monkeys}. Looped Transformers accumulate computation by
repeatedly applying a shared transformation to a latent state
\citep{dehghani2019universal,giannou2023looped,yang2024looped}. This extends computational depth
without adding parameters, and greater recurrent depth can improve reasoning performance
\citep{saunshi2025latent,geiping2025huginn}. Making these models effective therefore involves
learning how successive updates should contribute to solving a problem.

The learning objective can reflect how predictions are used at inference. Best-of-$N$ sampling
generates and verifies multiple candidate solutions, increasing the opportunity to find a correct
answer beyond a single attempt \citep{brown2024monkeys,cobbe2021verifiers}. Inference-aware
Best-of-$N$ training incorporates this procedure into learning \citep{chow2025inference}, while
Explorative Modeling (XM) trains on the candidate that best matches a target
\citep{gladstone2026explorative}. These approaches illustrate how learning objectives shape
candidate generation. Recurrent models also offer multiple predictions, read out at successive
depths along one trajectory. We therefore ask: \emph{Can a single recurrent trajectory be trained
to visit distinct useful solution candidates?}

Simply increasing recurrent computation does not guarantee distinct useful candidates. Recent
work promotes fixed-point convergence for stability \citep{yang2026stabilizingrecurrentdynamicstesttime}
or uses convergence for halting \citep{movahedi2026fixedpointreasonersstableadaptive}, but convergence
alone does not determine which alternatives a trajectory discovers. Supervising only the final
prediction \citep{dehghani2019universal,saunshi2025latent} does not directly reward useful intermediate
alternatives; supervising every depth against the same target asks all readouts to reproduce that
target. Candidate-set evaluation suggests another criterion: the target should be reached somewhere
within the depth budget. Because later predictions depend on preceding latent states, the learning
problem is to train these coupled predictions collectively.

% Keep Figure 1 with its caption in the text flow at the top of page 2.
\noindent\begin{minipage}{\linewidth}
  \centering
  \includegraphics[width=0.85\linewidth]{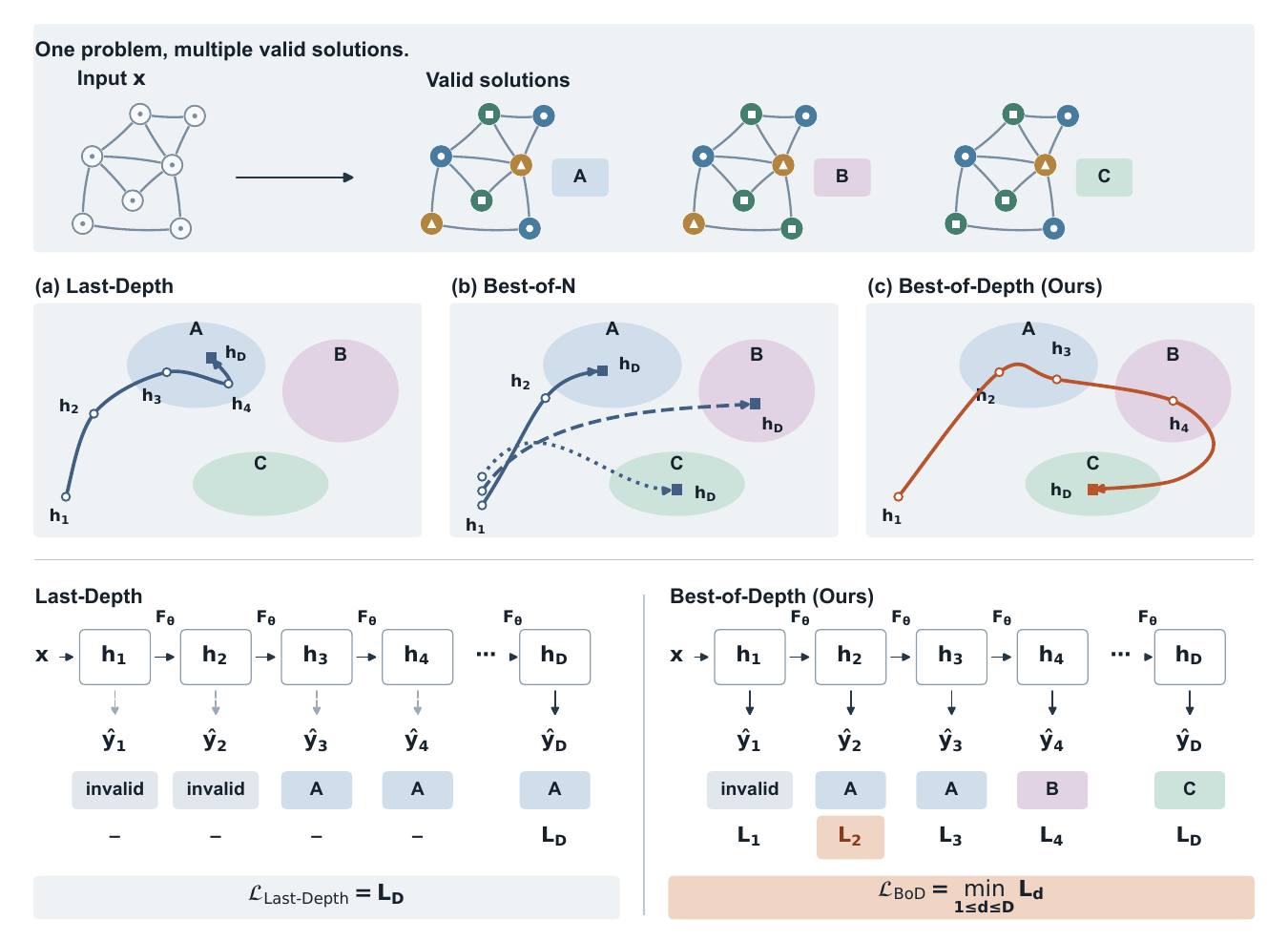}
  \captionof{figure}{\textbf{Learning alternative solutions along one recurrent trajectory.}
    \textbf{Top:} One input admits multiple valid solutions. \textbf{Middle:} Schematic latent
    trajectories illustrate (a)~repeated predictions under Last-Depth, (b)~independent candidates
    from Best-of-$N$, and (c)~distinct solutions along one BoD trajectory; shaded regions decode
    to A--C. \textbf{Bottom:} Both models reuse $F_\theta$ across depths. Last-Depth trains on
    $L_D$; BoD trains on the lowest loss across depths (highlighted $L_2$ for target A).}
  \label{fig:overview}
\end{minipage}

We introduce \textbf{Best-of-Depth (\ourmethod{})}, which trains a recurrent trajectory as a
candidate set using the minimum prediction loss across supervised depths (Figure~\ref{fig:overview}).
A shared decoder reads out predictions, and the loss at the selected depth trains the recurrent
computation. Each target need only be matched at one depth, allowing different valid targets for
the same input to be matched at different depths without a prescribed assignment or visitation
order. The set-loss formulation is related to multiple-hypothesis learning
\citep{guzmanrivera2012mcl,lee2016smcl,bhattacharyya2018bestofmany}; here, candidates are generated
successively and coupled through shared latent dynamics. BoD requires no architectural changes,
additional parameters, or depth-specific labels.

We study solution discovery theoretically and empirically. Under exact-match loss and uniformly
sampled valid targets, minimizing expected BoD loss maximizes solution coverage. On N-Queens
completion and graph coloring, where all valid solutions can be enumerated, Success@Depth measures
whether any valid solution is found, while Coverage@Depth measures the fraction of the valid
solution set recovered. Coverage therefore tests whether additional computation discovers alternatives
after an initial solution is found. We also test discovery beyond teacher pools and training depth,
and compare trajectory counts and depths under a fixed recurrent-step budget on these controlled
tasks. For autoregressive code generation, we fine-tune Huginn \citep{geiping2025huginn} and evaluate
programs generated separately at different fixed depths.

We find that BoD recovers more distinct valid solutions on both controlled tasks.
The N-Queens coverage reaches 91.03%, versus 45.75% for final-depth training. On graph coloring,
coverage continues increasing after success nearly saturates, showing further discovery of valid
alternatives. BoD also recovers solutions withheld from teacher pools on familiar inputs and finds
new solutions beyond the training depth. BoD achieves higher pooled code-generation success than
final-depth training on HumanEval and MBPP \citep{chen2021codex,austin2021mbpp}. On MBPP, it also
gains more solved problems from pooling beyond depth 32 alone. Fixed-budget comparisons favor combining depth with
multiple trajectories. Together with analyses of outputs and latent states, these results characterize
how candidate-set training changes solution discovery along recurrent trajectories.

% =============================================================================
\section{Related Work}
\label{sec:related}

\paragraph{Recurrent Computation and Looped Transformers.}
\label{sec:related-recurrence}
Looped Transformers reuse layers across depth; the Universal Transformer also supports adaptive
halting \citep{dehghani2019universal}. Subsequent work studies algorithmic capabilities
\citep{giannou2023looped,yang2024looped,gatmiry2024multistep} and latent reasoning
\citep{saunshi2025latent}. At language-model scale, \citet{geiping2025huginn} study reasoning gains
from increased recurrent depth. Flexible depth allocation uses learned exits
\citep{zhu2026scalinglatentreasoninglooped}, token-level routing \citep{bae2025mixture}, or shortcut
consistency \citep{jeddi2026loopformer}. Stable depth extrapolation motivates constraints on recurrent
dynamics and depth sampling \citep{prairie2026parcaescalinglawsstable}, or depth-dependent residual
scaling \citep{li2026deeploop}.

Existing methods supervise final outputs at fixed depths \citep{dehghani2019universal,saunshi2025latent}
or sampled depths \citep{geiping2025huginn}, apply supervision over recurrent windows
\citep{yang2024looped}, or supervise intermediate outputs for early exit
\citep{bae2025relaxedrecursivetransformerseffective}. Other objectives weight losses by exit
probabilities \citep{zhu2026scalinglatentreasoninglooped}. Unlike intermediate reasoning-token supervision
\citep{fan2026lotus}, BoD needs only output targets and prescribes no solution order.

\paragraph{Test-Time Candidate Generation and Compute Scaling.}
\label{sec:related-inference}
Additional inference compute supports candidate generation and evaluation. Repeated sampling
raises the chance of at least one correct candidate, while identifying it remains a separate problem
\citep{brown2024monkeys}. Best-of-$N$ selects candidates with a verifier \citep{cobbe2021verifiers}; self-consistency
aggregates answers across reasoning paths \citep{wang2023selfconsistency}. Process supervision
assesses intermediate steps \citep{lightman2024verify}, while tree-structured methods evaluate
partial solutions to guide further expansion \citep{yao2023tot,snell2025scaling}.

\citet{snell2025scaling} compare parallel sampling with sequential revisions, finding that the
preferred allocation depends on difficulty and budget. Their revision process is conditioned on previous
textual answers. Our fixed recurrent-step comparison instead allocates compute between
recurrent latent updates and multiple trajectories (Section~\ref{sec:widthdepth}), with external
candidate evaluation.

\paragraph{Learning to Generate Multiple Hypotheses.}
\label{sec:related-hypotheses}
Multiple Choice Learning scores a prediction set through its best member's loss against the target
\citep{guzmanrivera2012mcl}. Stochastic MCL trains only the lowest-loss predictor per example,
enabling deep ensembles to specialize without a predefined data partition \citep{lee2016smcl}.
Best-of-Many uses the best reconstruction among latent samples in a variational objective for
sequence prediction \citep{bhattacharyya2018bestofmany}. The forward formulation of Explorative
Modeling learns from the candidate closest to a target, studying exploration as a scaling axis
across generative tasks \citep{gladstone2026explorative}.

\citet{chow2025inference} align imitation and reinforcement learning with verifier-based Best-of-$N$
selection. BoD instead minimizes supervised prediction loss over recurrent readouts: later
hypotheses depend on earlier states, and the selected loss trains the shared update and decoder.

% =============================================================================
\section{Best-of-Depth}
\label{sec:method}

\subsection{Looped Transformers}
\label{sec:looped}
A looped Transformer encodes an input $x$ and repeatedly applies a shared transformation:
\begin{equation}
 h_0=\Enc_\theta(x),\qquad h_d=F_\theta(h_{d-1};x),\qquad
 p_d=\Dec_\theta(h_d),\quad d=1,\ldots,D.
 \label{eq:loop}
\end{equation}
One application of $F_\theta$ is a \emph{loop}, and $d$ denotes \emph{recurrent depth}. A shared
decoder reads the prediction distribution $p_d$ from each state. Let $y\in\mathcal Y^\star(x)$ be
a valid target, $\hat y_d(x)$ the greedy prediction, and $L_d(x,y)=\ell(p_d,y)$ the per-example loss.
We omit input and parameter dependence when unambiguous.

Supervising the final recurrent output is an established approach in fixed-depth looped models
\citep{dehghani2019universal,saunshi2025latent}. We call this \emph{Last-Depth training}: only the
prediction at the maximum training depth contributes to the loss,
\begin{equation}
 \cL_{\text{Last-Depth}}=L_D.
 \label{eq:standard}
\end{equation}
\subsection{Best-of-Depth Objective}
\label{sec:objective}
At training depth $D=\Dtr$, let $\mathcal D\subseteq\{1,\ldots,D\}$ denote the supervised
depths, including $D$. BoD minimizes
\begin{equation}
 \cL_{\text{BoD}}(x,y)=\min_{d\in\mathcal D}L_d(x,y).
 \label{eq:bod}
\end{equation}
Each $L_d$ averages losses over predicted positions. BoD selects one depth for the entire target
and trains on its loss, breaking exact ties at random.

For an autoregressive target, the teacher-forced loss is
\begin{equation}
 L_d(x,y)=-\frac{1}{T}\sum_{t\in\mathcal T}
 \log p_d(y_t\mid x,y_{<t}),\qquad T=|\mathcal T|,
 \label{eq:sequence}
\end{equation}
where $\mathcal T$ contains target-token positions and $p_d$ is the depth-$d$ next-token distribution.

For the same readouts, Mean-Depth supervises all depths and Random-Depth samples one per example:
\begin{equation}
 \cL_{\text{Mean-Depth}}=\frac{1}{|\mathcal D|}\sum_{d\in\mathcal D}L_d,\qquad
 \cL_{\text{Random-Depth}}=L_{\tilde d},\quad\tilde d\sim\mathrm{Uniform}(\mathcal D).
 \label{eq:baselines}
\end{equation}
Soft-Depth smoothly relaxes the minimum:
\begin{equation}
 \cL_{\text{Soft-Depth}}(x,y)
 =-\tau\log\left(\frac{1}{|\mathcal D|}\sum_{d\in\mathcal D}
 \exp\!\left[-L_d(x,y)/\tau\right]\right),\qquad \tau>0.
 \label{eq:softdepth}
\end{equation}
Lower-loss depths receive greater gradient weight; the limits $\tau\to0$ and $\tau\to\infty$
recover BoD and Mean-Depth, respectively. All methods allow the same readouts at inference time.

\subsection{Interpreting the Best-of-Depth Objective}
\label{sec:interp}

For exact-match loss, write
$\cL_{\mathrm{BoD}}^{0\text{--}1}=\min_{d\in\mathcal D}\1[\hat y_d\ne y]$.
The first two results connect this objective to target discovery and solution coverage.

\begin{theorybox}
\begin{proposition}[First-hit interpretation]
\label{prop:firsthit}
Let $\tau_y=\min\{d\in\mathcal D:\hat y_d=y\}$, with $\tau_y=\infty$ if no readout matches $y$.
Then
\begin{equation}
\label{eq:firsthit}
\mathbb E[\cL_{\mathrm{BoD}}^{0\text{--}1}]=\Pr(\tau_y>D).
\end{equation}
\end{proposition}
\end{theorybox}
BoD thus evaluates whether the target is reached anywhere within the depth budget.

\begin{theorybox}
\begin{proposition}[Connection to solution coverage]
\label{prop:coverage}
Fix $x$ and $S_D(x)=\{\hat y_d(x):d\in\mathcal D\}$ independently of the target
$Y\sim\mathrm{Uniform}(\mathcal Y^\star(x))$. Then
\begin{equation}
\label{eq:coverage-risk}
\mathbb E_Y[\cL_{\mathrm{BoD}}^{0\text{--}1}\mid x]
=1-\frac{|S_D(x)\cap\mathcal Y^\star(x)|}{|\mathcal Y^\star(x)|}
=1-\mathrm{Coverage@Depth}(D;x).
\end{equation}
\end{proposition}
\end{theorybox}
Minimizing this risk maximizes coverage; repeated candidates do not increase coverage.

For cross-entropy, the following construction separates the objectives under shared recurrent
computation.
\begin{theorybox}
\begin{proposition}[Separation in solution discovery]
\label{prop:separation}
There exist a multi-solution task, a shared recurrent model with depth budget $D$ and factorized readouts, and a
common symmetric initialization and SGD schedule such that, under cross-entropy training with
uniformly resampled valid targets, greedy coverage satisfies
\begin{equation}
\label{eq:objective-separation}
\lim_{n\to\infty}\operatorname{Coverage@Depth}(D;\theta_A^{(n)})=
\begin{cases}
0,& A\in\{\text{Last, Mean, Random, Soft}\},\\
1,& A=\text{BoD},
\end{cases}
\end{equation}
almost surely, where $\theta_A^{(n)}$ denotes the parameters after $n$ updates with objective $A$.
The Soft-Depth result holds for every fixed $\tau>0$; BoD selects one loss-minimizing depth when losses tie.
\end{proposition}
\end{theorybox}
In this prescribed cyclic recurrence, the other objectives retain or recover a symmetric predictor
with invalid greedy outputs. BoD breaks symmetry by selecting one depth, allowing different targets
to reinforce the same latent solution through different recurrent phases.

Proofs are in Appendices~\ref{app:exact-match-proof} and~\ref{app:separation-proof}, including a
numerical illustration of Proposition~\ref{prop:separation}. Cross-entropy coverage bounds are
given in Appendices~\ref{app:discovery-proof} and~\ref{app:greedy-proof}.

% =============================================================================

\section{Experiments}
\label{sec:exp}

\subsection{Experimental Setup}
\label{sec:setup}

\paragraph{Compared methods and candidate generation.}
We compare BoD, Last-Depth, Mean-Depth, Random-Depth, and Soft-Depth ($\tau=0.25$;
Section~\ref{sec:method}). Within each task, all share the output head and a width-128 Transformer
with two shared layers per loop and eight attention heads.
Unless stated otherwise, we train with uniformly resampled valid targets at $\Dtr=16$ and
evaluate greedy readouts through $\Dte=64$. For controlled-task experiments, we report means
and sample standard deviations (SD) over three training seeds; randomized evaluations first average
inference repeats within each seed.
For multiple trajectories, we pool all intermediate predictions. Section~\ref{sec:code}
describes autoregressive generation; Appendix~\ref{app:impl} gives implementation details.

\paragraph{Evaluation metrics.}
For the set of distinct candidates $S_D(x)$ (Section~\ref{sec:interp}), with $D=\Dte$, we report
\begin{align}
 \text{Success@Depth}(D;x)
 &=\1[S_D(x)\cap\mathcal Y^\star(x)\ne\varnothing],\\
 \text{Coverage@Depth}(D;x)
 &=\frac{|S_D(x)\cap\mathcal Y^\star(x)|}{|\mathcal Y^\star(x)|}.
\end{align}
Both metrics are percentages averaged over problems. Coverage is our primary controlled-task
metric; Success requires only a validity verifier. Single-depth accuracy measures the fraction
of problems solved at depth $D$. We abbreviate cumulative metrics as Success@$D$ and Coverage@$D$,
or Success and Coverage when candidate sets have no common depth budget.

\begin{figure}[t]
  \centering
  \includegraphics[width=0.95\linewidth]{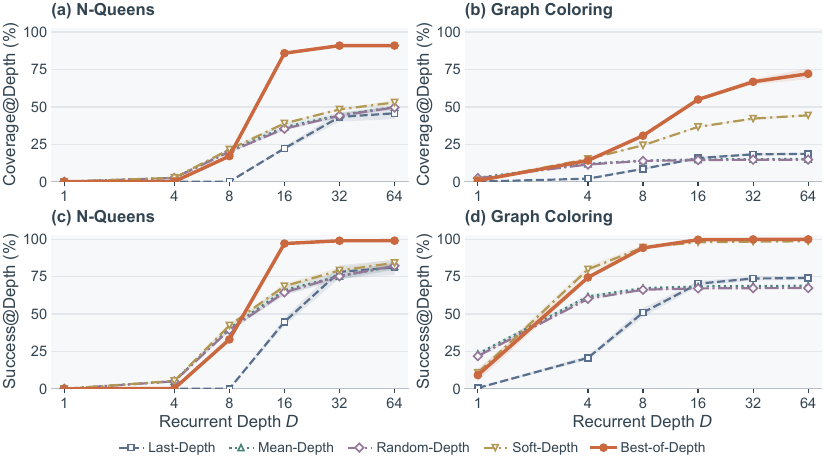}
  \caption{\textbf{Depth-wise solution discovery.} Coverage@Depth (top) and Success@Depth
    (bottom) for greedy trajectories trained at depth 16. Bands indicate $\pm1$ SD.}
  \label{fig:depth}
\end{figure}

\subsection{N-Queens Completion}
\label{sec:nqueens}

\noindent\textbf{Settings.} Each input specifies five fixed queens on a $10\times10$ board. The model
completes it so no two queens share a row, column, or diagonal. We use 100 cell tokens and
train on 24,320 problems. Evaluation uses 1,000 multi-solution test problems with all valid
completions enumerated. Distinct completed boards are distinct solutions; symmetry-equivalent
inputs remain in the same split
(Appendix~\ref{app:nqueens}).

\noindent\textbf{Results.} BoD outperforms all four objectives at depth 64, with 91.03\% coverage
and 99.03\% success (Figure~\ref{fig:depth}, left). Its largest gains occur near the training
depth, with further discoveries beyond it and saturation by depth 64. Deeper evaluation benefits
all objectives, but the others recover fewer valid completions per problem. BoD thus improves
both success and the fraction of alternatives recovered along one trajectory.

\subsection{Graph Coloring}
\label{sec:coloring}

\noindent\textbf{Settings.} The model assigns one of three colors to each vertex so that
adjacent vertices have different colors. We train on 4,096 connected 10-vertex graphs and test
on 512, with adjacency rows encoded as vertex tokens. Each graph has 2--128 enumerated
solution families; colorings that differ only in color names count as one solution.
Isomorphic graphs do not cross splits (Appendix~\ref{app:coloring}).

\noindent\textbf{Results.} BoD achieves 72.16\% coverage at depth 64 and solves nearly all test graphs
(Figure~\ref{fig:depth}, right). Success nearly saturates at the training depth, while coverage
continues increasing: additional depth mainly recovers alternatives for already-solved graphs.
Soft-Depth achieves high success but only 44.39\% coverage, revealing differences in solution discovery
despite similar success. Its coverage remains below BoD after development-set temperature
selection (Appendix~\ref{app:temperature-results}).

\subsection{Code Generation}
\label{sec:code}

\noindent\textbf{Settings.} We fine-tune Huginn, a pretrained 3B-parameter looped language
model \citep{geiping2025huginn}, on 20,000 Python examples from CodeFeedback
\citep{zheng2024opencodeinterpreter}, with 1,000 validation examples. We train models under all five objectives
at depth 32; we also evaluate the pretrained model. On all 164 HumanEval and 500 MBPP
problems \citep{chen2021codex,austin2021mbpp}, each model generates a complete program separately
at each fixed depth in $\{4,8,12,16,24,32\}$ using greedy decoding. Success@32 pools these six
candidates. We score completed programs using tests that do not guide generation or repair; MBPP uses
the public tests included in the prompt. Appendix~\ref{app:code} gives the full protocol.

\noindent\textbf{Results.} BoD achieves the highest Success@32 on both benchmarks (Table~\ref{tab:code}):
37.8\% versus Last-Depth's 34.8\% on HumanEval, and 44.2\% versus 39.2\% on MBPP. BoD also has
the highest depth-32 accuracy, although Mean-Depth, Random-Depth, and Soft-Depth perform better
at shallow depths. Pooling BoD's depth-specific programs solves problems missed at depth 32 alone,
demonstrating useful shallow-depth candidates and extending BoD's benefit to autoregressive generation.

\begin{table}[!htbp]
\centering
\caption{Code-generation results (\%). Base is pretrained; other rows denote fine-tuning objectives. Depth columns show accuracy for one greedy program per problem. Success@32 pools the six evaluated depths $\{4,8,12,16,24,32\}$. Bold marks each column's best result.}
\label{tab:code}
\small
\setlength{\tabcolsep}{3.3pt}
\renewcommand{\arraystretch}{1.14}
\begin{tabularx}{\linewidth}{@{}>{\raggedright\arraybackslash}p{65pt}*{4}{>{\centering\arraybackslash}X}>{\centering\arraybackslash}p{60pt}*{4}{>{\centering\arraybackslash}X}>{\centering\arraybackslash}p{60pt}@{}}
\toprule
 & \multicolumn{5}{c}{\textbf{HumanEval}} & \multicolumn{5}{c}{\textbf{MBPP}} \\
\cmidrule(lr){2-6}\cmidrule(lr){7-11}
 & \multicolumn{4}{c}{Depth} & Success@Depth & \multicolumn{4}{c}{Depth} & Success@Depth \\
\cmidrule(lr){2-5}\cmidrule(lr){6-6}\cmidrule(lr){7-10}\cmidrule(lr){11-11}
Method & 4 & 8 & 16 & 32 & 32 & 4 & 8 & 16 & 32 & 32 \\
\midrule
Base & 0.0 & 12.8 & 24.4 & 25.6 & 30.5 & 1.2 & 17.0 & 29.0 & 29.8 & 39.8 \\
Last-Depth & 0.0 & 13.4 & 27.4 & 26.8 & 34.8 & 1.8 & 21.2 & 30.2 & 31.4 & 39.2 \\
Mean-Depth & \textbf{6.7} & 20.1 & 27.4 & 27.4 & 31.7 & \textbf{10.4} & \textbf{26.0} & 28.0 & 29.0 & 37.2 \\
Random-Depth & 4.9 & 20.1 & 26.8 & 28.7 & 30.5 & 8.8 & 25.8 & 29.6 & 29.6 & 37.4 \\
Soft-Depth & 6.1 & \textbf{20.7} & 26.2 & 26.8 & 32.9 & 9.4 & 25.2 & 29.2 & 30.8 & 39.4 \\
\rowcolor{oursbg} \textbf{Best-of-Depth} & 0.6 & 14.6 & \textbf{29.3} & \textbf{29.9} & \textbf{37.8} & 0.6 & 21.0 & \textbf{30.6} & \textbf{33.2} & \textbf{44.2} \\
\bottomrule
\end{tabularx}
\end{table}

\subsection{Ablation Studies}
\label{sec:ablation}

\begin{figure}[!t]
  \centering
  \includegraphics[width=0.95\linewidth]{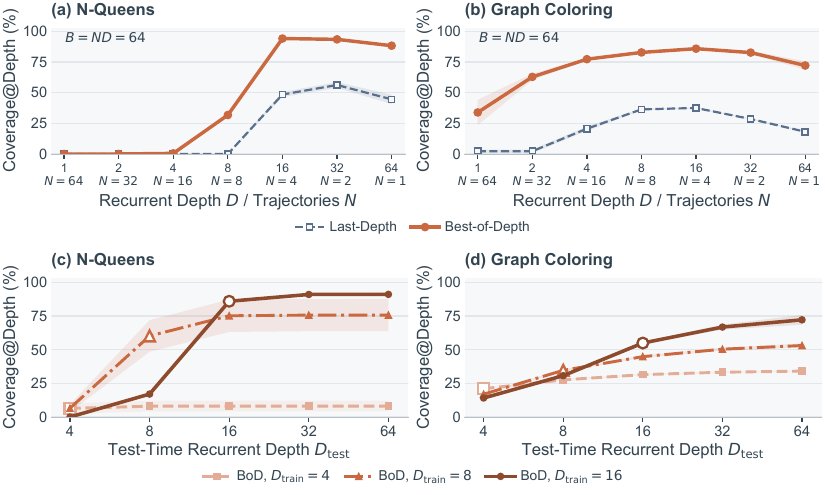}
  \caption{\textbf{Depth allocation and extrapolation.} (a, b)~Coverage pooling intermediate
    predictions from randomized trajectories under $ND=64$. (c, d)~BoD with unperturbed greedy
    readouts; open markers denote training depth. Bands indicate $\pm1$ SD.}
  \label{fig:compute-depth}
  \label{fig:width}
  \label{fig:train}
\end{figure}

\subsubsection{Width vs. Depth under a Fixed Recurrent-Step Budget}
\label{sec:widthdepth}

Figure~\ref{fig:compute-depth}(a, b) compares seven allocations of $B=ND=64$ recurrent-block
applications, with $D\in\{1,2,4,8,16,32,64\}$ and $N=64/D$. Each method pools and deduplicates
all intermediate predictions from $N$ trajectories. We perturb the initial states with Gaussian noise
for N-Queens and permute vertices for graph coloring, including at $N=1$.
Appendix~\ref{app:full} gives the protocols.

BoD achieves its highest mean coverage at $(N,D)=(4,16)$ on both tasks; Last-Depth peaks at
$(2,32)$ on N-Queens and $(4,16)$ on graph coloring. These results favor combining depth with
multiple trajectories.

\subsubsection{Training Depth and Test-Time Extrapolation}
\label{sec:traindepth}

We vary training depth to examine whether solution discovery continues beyond the supervised
trajectory. Figure~\ref{fig:compute-depth}(c, d) evaluates BoD trained at $\Dtr\in\{4,8,16\}$
using greedy readouts through $\Dte\in\{4,8,16,32,64\}$. All settings show further discoveries
beyond the training depth, with diminishing gains. Models trained deeper reach higher coverage
by depth 64, but can lag at shallower depths. Training depth thus shapes both the breadth of the
candidate set and the depth at which alternatives become available. Success curves and numerical
results are in Appendix~\ref{app:full}.

\subsubsection{Discovering Solutions Beyond the Training Targets}
\label{sec:unseen-solutions}

We test discovery beyond supplied targets by restricting each graph's teacher pool to one, two,
three, or four valid solutions, with an all-solutions control. The pools are nested, and targets are sampled
uniformly from each pool; all conditions use the same training budget and 3,034 training graphs with at least
five solutions (Appendix~\ref{app:teacher-diversity}). Evaluation on these same graphs measures
\emph{Seen Coverage}, the fraction of teacher-pool solutions recovered, and \emph{Unseen Coverage},
the fraction of withheld valid solutions recovered, averaged per graph.
Table~\ref{tab:teacher-diversity} compares Last-Depth, Last-Depth with multiple sampled trajectories
(Last-Depth + BoN), and BoD under a recurrent-step budget of 64, collecting intermediate predictions
for all three. BoD has the highest unseen coverage in every restricted-pool condition, increasing
from 29.4\% with one training solution to 46.2\% with four. These are familiar training inputs, not unseen graphs.

% Declare the analysis float here so it stays near Section 5.
\begin{figure}[!t]
  \centering
  \includegraphics[width=0.95\linewidth]{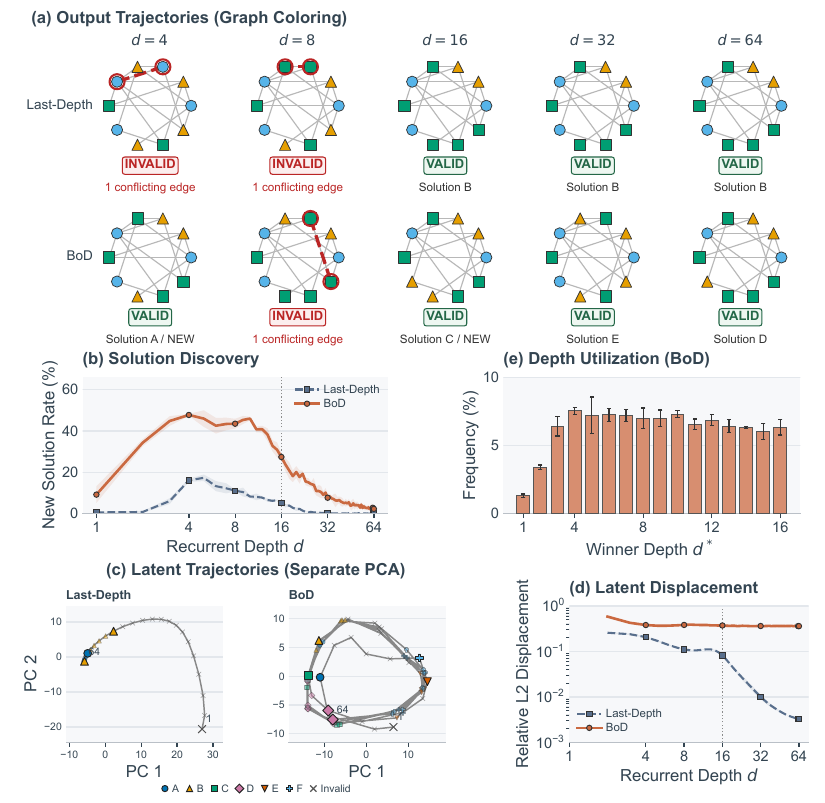}
  \caption{\textbf{Exploratory recurrent dynamics on graph coloring.}
    (a)~Example outputs; NEW marks first appearances. (b)~Fraction of graphs with a new valid
    solution. (c)~Separate PCA projections for the same example and seed as (a); markers
    identify solutions or invalid outputs. (d)~Relative state displacement. (e)~Loss-minimizing
    depths. Bands (b, d) and error bars (e) indicate $\pm1$ SD. Dotted lines mark training depth 16.}
  \label{fig:analysis}
\end{figure}

\begin{table}[!htbp]
\centering
\caption{Discovery beyond the training targets on graph coloring. Seen Coverage and Unseen Coverage (\%) on 3,034 training graphs under $ND=64$. All uses every valid solution as a training target candidate; Unseen Coverage is then undefined (N/A).}
\label{tab:teacher-diversity}
\small
\setlength{\tabcolsep}{3pt}
\renewcommand{\arraystretch}{1.12}
\begin{tabularx}{\linewidth}{@{}c*{6}{>{\centering\arraybackslash}X}@{}}
\toprule
 & \multicolumn{2}{c}{Last-Depth} & \multicolumn{2}{c}{Last-Depth + BoN} & \multicolumn{2}{c}{Best-of-Depth} \\
\cmidrule(lr){2-3}\cmidrule(lr){4-5}\cmidrule(l){6-7}
\shortstack{Training\\solutions} & \shortstack{Seen\\Coverage} & \shortstack{Unseen\\Coverage} & \shortstack{Seen\\Coverage} & \shortstack{Unseen\\Coverage} & \shortstack{Seen\\Coverage} & \shortstack{Unseen\\Coverage} \\
\midrule
1 & $\mathbf{99.8}\!\pm\!0.0$ & $13.5\!\pm\!2.4$ & $25.4\!\pm\!0.3$ & $25.4\!\pm\!0.4$ & $98.8\!\pm\!0.4$ & $\mathbf{29.4}\!\pm\!0.4$ \\
2 & $46.6\!\pm\!0.6$ & $12.9\!\pm\!1.2$ & $22.8\!\pm\!0.7$ & $22.9\!\pm\!0.9$ & $\mathbf{94.2}\!\pm\!1.1$ & $\mathbf{37.0}\!\pm\!0.8$ \\
3 & $37.1\!\pm\!1.4$ & $13.8\!\pm\!1.3$ & $23.6\!\pm\!0.2$ & $23.7\!\pm\!0.3$ & $\mathbf{91.3}\!\pm\!1.3$ & $\mathbf{42.2}\!\pm\!0.5$ \\
4 & $28.4\!\pm\!0.3$ & $13.8\!\pm\!0.4$ & $24.4\!\pm\!0.6$ & $24.4\!\pm\!0.6$ & $\mathbf{87.7}\!\pm\!1.6$ & $\mathbf{46.2}\!\pm\!1.2$ \\
All & $11.8\!\pm\!0.2$ & N/A & $25.0\!\pm\!0.4$ & N/A & $\mathbf{67.4}\!\pm\!4.4$ & N/A \\
\bottomrule
\end{tabularx}
\end{table}

\subsubsection{Comparison with Other Baselines}
\label{sec:other-baselines}

Table~\ref{tab:other-baselines} in Appendix~\ref{app:other-baselines} compares BoD with unshared
and autoregressive Transformers,
masked diffusion~\citep{sahoo2024simpleeffectivemaskeddiffusion},
HRM~\citep{wang2025hierarchicalreasoningmodel},
TRM~\citep{jolicoeurmartineau2025morerecursivereasoningtiny}, and an instance-level adaptation of
Ouro's objective~\citep{zhu2026scalinglatentreasoninglooped}. Task splits and update counts are
shared, but architectures and candidate counts vary. BoD
achieves the highest mean success and coverage on both tasks. HRM has similar N-Queens success
but substantially lower coverage, illustrating the distinction between finding one solution and
many. Ouro's adapted objective also yields lower coverage on the same looped backbone.

% =============================================================================

\section{Understanding Exploratory Recurrent Dynamics}
\label{sec:analysis}

Figure~\ref{fig:analysis} relates BoD's broader discovery to its recurrent dynamics on graph coloring.
The statistics cover all 512 test graphs and three training seeds (Appendix~\ref{app:analysis-protocol}).

BoD discovers new valid solutions near and just beyond the training depth, where Last-Depth
makes few discoveries (Figure~\ref{fig:analysis}b). This metric counts first solutions and subsequent
alternatives. Discovery declines at greater depths (Figure~\ref{fig:depth}). In panel (a), Last-Depth
produces invalid outputs through depth 14 and reaches two distinct solutions by depth 64,
whereas BoD reaches all six.

Separate PCA projections for this example show BoD moving among states decoding to different
valid solutions (Figure~\ref{fig:analysis}c). Across the test set, BoD maintains greater relative
state displacement near and beyond the training depth, while Last-Depth's declines (panel (d)). Displacement is measured
before projection relative to the preceding state's norm. Panels (a, b) measure discovery;
panels (c, d) characterize the accompanying dynamics.

Loss-minimizing depths span all 16 supervised depths (Figure~\ref{fig:analysis}e), providing the best
predictions for different input--target pairs. The histogram weights graphs equally and valid targets
uniformly within each graph. Together, these analyses associate broader discovery with sustained
latent movement and useful intermediate readouts. Additional computation beyond the training
depth yields further alternatives.

% =============================================================================

\section{Discussion and Limitations}
\label{sec:discussion}

Our results highlight training objectives as a design choice for inference-time candidate
generation. On controlled tasks, BoD improves coverage using the same architecture and inference
readouts as the comparison objectives. This motivates aligning recurrent training with candidate-set
evaluation so intermediate predictions contribute useful alternatives.

Fixed-budget results favor combining depth with multiple trajectories. For autoregressive generation,
extensions include cross-depth sampling, branching, restarts, verification, and adaptive stopping.

Our evidence covers small combinatorial tasks and one pretrained language model; discovery
saturates with depth. Generality requires testing larger tasks and additional models.

% =============================================================================
\section{Conclusion}
\label{sec:conclusion}
Best-of-Depth trains on the lowest-loss prediction along a looped Transformer's trajectory,
allowing different depths to match different valid targets without architectural changes.
On N-Queens completion and graph coloring, it improves success and solution coverage,
with further discoveries beyond the training depth. On familiar graph-coloring inputs, it also
recovers solutions withheld from the teacher pool. Trajectory analyses show broader discovery
alongside sustained latent movement and use of multiple intermediate depths. Under a fixed
recurrent-step budget, combining depth with multiple trajectories yields higher coverage.
Code-generation experiments show higher cumulative success when pooling programs from different
fixed depths. These findings support training and evaluating recurrent trajectories as candidate
sets of complementary valid solutions across depth.

\clearpage
\section*{Reproducibility Statement}
Appendix~\ref{app:impl} describes the datasets, models, training objectives, and
evaluation protocols. Appendix~\ref{app:analysis-protocol} specifies the trajectory analyses, Appendix~\ref{app:full} reports detailed results and aggregation
procedures, and Appendix~\ref{app:theory} provides the proofs. Table~\ref{tab:huginn} lists the code-generation hyperparameters used for all five
fine-tuning objectives.

\section*{AI Use Statement}
AI tools assisted with manuscript editing, mathematical derivations and proof review, LaTeX
formatting, and compilation.

\bibliographystyle{iclr2027_conference}
\bibliography{refs}

@inproceedings{geiping2025huginn,
  title     = {Scaling up Test-Time Compute with Latent Reasoning: A Recurrent Depth Approach},
  author    = {Geiping, Jonas and McLeish, Sean and Jain, Neel and Kirchenbauer, John and Singh, Siddharth and Bartoldson, Brian R. and Kailkhura, Bhavya and Bhatele, Abhinav and Goldstein, Tom},
  booktitle = {Advances in Neural Information Processing Systems},
  year      = {2025},
  url = {https://openreview.net/forum?id=S3GhJooWIC}
}

@inproceedings{snell2025scaling,
  title     = {Scaling {LLM} Test-Time Compute Optimally Can be More Effective than Scaling Parameters for Reasoning},
  author    = {Snell, Charlie and Lee, Jaehoon and Xu, Kelvin and Kumar, Aviral},
  booktitle = {International Conference on Learning Representations},
  year      = {2025},
  url = {https://openreview.net/forum?id=4FWAwZtd2n}
}

@inproceedings{chow2025inference,
  title     = {Inference-Aware Fine-Tuning for Best-of-{N} Sampling in Large Language Models},
  author    = {Chow, Yinlam and Tennenholtz, Guy and Gur, Izzeddin and Zhuang, Vincent and Dai, Bo and Thiagarajan, Sridhar and Boutilier, Craig and Agarwal, Rishabh and Kumar, Aviral and Faust, Aleksandra},
  booktitle = {International Conference on Learning Representations},
  year      = {2025},
  url = {https://arxiv.org/abs/2412.15287}
}

@inproceedings{guzmanrivera2012mcl,
  title     = {Multiple Choice Learning: Learning to Produce Multiple Structured Outputs},
  author    = {Guzm{\'a}n-Rivera, Abner and Batra, Dhruv and Kohli, Pushmeet},
  booktitle = {Advances in Neural Information Processing Systems},
  year      = {2012},
  url = {https://proceedings.neurips.cc/paper_files/paper/2012/hash/cfbce4c1d7c425baf21d6b6f2babe6be-Abstract.html}
}

@inproceedings{lee2016smcl,
  title     = {Stochastic Multiple Choice Learning for Training Diverse Deep Ensembles},
  author    = {Lee, Stefan and Purushwalkam, Senthil and Cogswell, Michael and Ranjan, Viresh and Crandall, David and Batra, Dhruv},
  booktitle = {Advances in Neural Information Processing Systems},
  year      = {2016},
  url = {https://arxiv.org/abs/1606.07839}
}

@inproceedings{bhattacharyya2018bestofmany,
  title     = {Accurate and Diverse Sampling of Sequences Based on a ``Best of Many'' Sample Objective},
  author    = {Bhattacharyya, Apratim and Schiele, Bernt and Fritz, Mario},
  booktitle = {Proceedings of the IEEE Conference on Computer Vision and Pattern Recognition},
  year      = {2018},
  url = {https://arxiv.org/abs/1806.07772}
}

@inproceedings{dehghani2019universal,
  title     = {Universal Transformers},
  author    = {Dehghani, Mostafa and Gouws, Stephan and Vinyals, Oriol and Uszkoreit, Jakob and Kaiser, {\L}ukasz},
  booktitle = {International Conference on Learning Representations},
  year      = {2019},
  url = {https://arxiv.org/abs/1807.03819}
}

@misc{gladstone2026explorative,
      title={Explorative Modeling: Unlocking a Third Pretraining Axis and End-to-End Generation},
      author={Alexi Gladstone and Heng Ji and Yilun Du},
      year={2026},
      eprint={2607.27372},
      archivePrefix={arXiv},
      primaryClass={cs.LG},
      url={https://arxiv.org/abs/2607.27372},
  howpublished = {arXiv preprint arXiv:2607.27372}
}

@misc{li2026deeploop,
      title={{DeepLoop}: Depth Scaling for Looped Transformers},
      author={Shuzhen Li and Yifan Zhang and Jiacheng Guo and Quanquan Gu and Mengdi Wang},
      year={2026},
      eprint={2607.13491},
      archivePrefix={arXiv},
      primaryClass={cs.LG},
      url={https://arxiv.org/abs/2607.13491},
  howpublished = {arXiv preprint arXiv:2607.13491}
}

@inproceedings{yang2026stabilizingrecurrentdynamicstesttime,
      title={Stabilizing Recurrent Dynamics for Test-Time Scalable Latent Reasoning in Looped Language Models},
      author={Xiao-Wen Yang and Ziyu Han and Xi-Hua Zhang and Wen-Da Wei and Jie-Jing Shao and Lan-Zhe Guo and Yu-Feng Li},
      year={2026},
      eprint={2605.26733},
      archivePrefix={arXiv},
      primaryClass={cs.LG},
      url={https://openreview.net/forum?id=tbdSfBMlbq},
  booktitle = {International Conference on Machine Learning}
}

@misc{movahedi2026fixedpointreasonersstableadaptive,
      title={Fixed-Point Reasoners: Stable and Adaptive Deep Looped Transformers},
      author={Sajad Movahedi and Vera Milovanovi{\'c} and Shlomo Libo Feigin and Alexander Theus and Thomas Hofmann and Valentina Boeva and T. Konstantin Rusch and Antonio Orvieto},
      year={2026},
      eprint={2606.18206},
      archivePrefix={arXiv},
      primaryClass={cs.AI},
      url={https://arxiv.org/abs/2606.18206},
  howpublished = {arXiv preprint arXiv:2606.18206}
}

@inproceedings{giannou2023looped,
  title     = {Looped Transformers as Programmable Computers},
  author    = {Giannou, Angeliki and Rajput, Shashank and Sohn, Jy-yong and Lee, Kangwook and Lee, Jason D. and Papailiopoulos, Dimitris},
  booktitle = {International Conference on Machine Learning},
  year      = {2023},
  url = {https://proceedings.mlr.press/v202/giannou23a.html}
}

@inproceedings{yang2024looped,
  title     = {Looped Transformers are Better at Learning Learning Algorithms},
  author    = {Yang, Liu and Lee, Kangwook and Nowak, Robert and Papailiopoulos, Dimitris},
  booktitle = {International Conference on Learning Representations},
  year      = {2024},
  url = {https://arxiv.org/abs/2311.12424}
}

@inproceedings{saunshi2025latent,
  title     = {Reasoning with Latent Thoughts: On the Power of Looped Transformers},
  author    = {Saunshi, Nikunj and Dikkala, Nishanth and Li, Zhiyuan and Kumar, Sanjiv and Reddi, Sashank J.},
  booktitle = {International Conference on Learning Representations},
  year      = {2025},
  url = {https://arxiv.org/abs/2502.17416}
}

@article{cobbe2021verifiers,
  title   = {Training Verifiers to Solve Math Word Problems},
  author  = {Cobbe, Karl and Kosaraju, Vineet and Bavarian, Mohammad and Chen, Mark and Jun, Heewoo and Kaiser, Lukasz and Plappert, Matthias and Tworek, Jerry and Hilton, Jacob and Nakano, Reiichiro and Hesse, Christopher and Schulman, John},
  journal = {arXiv preprint arXiv:2110.14168},
  year    = {2021},
  url = {https://arxiv.org/abs/2110.14168}
}

@article{brown2024monkeys,
  title   = {Large Language Monkeys: Scaling Inference Compute with Repeated Sampling},
  author  = {Brown, Bradley and Juravsky, Jordan and Ehrlich, Ryan and Clark, Ronald and Le, Quoc V. and R{\'e}, Christopher and Mirhoseini, Azalia},
  journal = {arXiv preprint arXiv:2407.21787},
  year    = {2024},
  url = {https://arxiv.org/abs/2407.21787}
}

@inproceedings{wang2023selfconsistency,
  title     = {Self-Consistency Improves Chain of Thought Reasoning in Language Models},
  author    = {Wang, Xuezhi and Wei, Jason and Schuurmans, Dale and Le, Quoc and Chi, Ed and Narang, Sharan and Chowdhery, Aakanksha and Zhou, Denny},
  booktitle = {International Conference on Learning Representations},
  year      = {2023},
  url = {https://arxiv.org/abs/2203.11171}
}

@inproceedings{lightman2024verify,
  title     = {Let's Verify Step by Step},
  author    = {Lightman, Hunter and Kosaraju, Vineet and Burda, Yura and Edwards, Harri and Baker, Bowen and Lee, Teddy and Leike, Jan and Schulman, John and Sutskever, Ilya and Cobbe, Karl},
  booktitle = {International Conference on Learning Representations},
  year      = {2024},
  url = {https://openreview.net/forum?id=v8L0pN6EOi}
}

@inproceedings{yao2023tot,
  title     = {Tree of Thoughts: Deliberate Problem Solving with Large Language Models},
  author    = {Yao, Shunyu and Yu, Dian and Zhao, Jeffrey and Shafran, Izhak and Griffiths, Thomas L. and Cao, Yuan and Narasimhan, Karthik},
  booktitle = {Advances in Neural Information Processing Systems},
  year      = {2023},
  url = {https://arxiv.org/abs/2305.10601}
}

@article{chen2021codex,
  title   = {Evaluating Large Language Models Trained on Code},
  author  = {Chen, Mark and Tworek, Jerry and Jun, Heewoo and Yuan, Qiming and Pinto, Henrique Ponde de Oliveira and Kaplan, Jared and others},
  journal = {arXiv preprint arXiv:2107.03374},
  year    = {2021},
  url = {https://arxiv.org/abs/2107.03374}
}

@article{austin2021mbpp,
  title   = {Program Synthesis with Large Language Models},
  author  = {Austin, Jacob and Odena, Augustus and Nye, Maxwell and Bosma, Maarten and Michalewski, Henryk and Dohan, David and Jiang, Ellen and Cai, Carrie and Terry, Michael and Le, Quoc and Sutton, Charles},
  journal = {arXiv preprint arXiv:2108.07732},
  year    = {2021},
  url = {https://arxiv.org/abs/2108.07732}
}

@misc{fan2026lotus,
      title={Bridging the Gap Between Latent and Explicit Reasoning with Looped Transformers},
      author={Ying Fan and Anej Svete and Kangwook Lee},
      year={2026},
      eprint={2606.31779},
      archivePrefix={arXiv},
      primaryClass={cs.LG},
      url={https://arxiv.org/abs/2606.31779},
  howpublished = {arXiv preprint arXiv:2606.31779}
}

@inproceedings{jeddi2026loopformer,
      title={{LoopFormer}: Elastic-Depth Looped Transformers for Latent Reasoning via Shortcut Modulation},
      author={Ahmadreza Jeddi and Marco Ciccone and Babak Taati},
      year={2026},
      booktitle={International Conference on Learning Representations},
      eprint={2602.11451},
      archivePrefix={arXiv},
      primaryClass={cs.CL},
      url={https://openreview.net/forum?id=RzYXb5YWBs},
}

@inproceedings{bae2025mixture,
      title={Mixture-of-Recursions: Learning Dynamic Recursive Depths for Adaptive Token-Level Computation},
      author={Sangmin Bae and Yujin Kim and Reza Bayat and Sungnyun Kim and Jiyoun Ha and Tal Schuster and Adam Fisch and Hrayr Harutyunyan and Ziwei Ji and Aaron Courville and Se-Young Yun},
      year={2025},
      eprint={2507.10524},
      archivePrefix={arXiv},
      primaryClass={cs.CL},
      url={https://openreview.net/forum?id=QuqsEIVWIG},
  booktitle = {Advances in Neural Information Processing Systems}
}

@InProceedings{gatmiry2024multistep,
  title = {Can Looped Transformers Learn to Implement Multi-step Gradient Descent for In-context Learning?},
  author = {Gatmiry, Khashayar and Saunshi, Nikunj and Reddi, Sashank J. and Jegelka, Stefanie and Kumar, Sanjiv},
  booktitle = {Proceedings of the 41st International Conference on Machine Learning},
  pages = {15130--15152},
  year = {2024},
  volume = {235},
  series = {Proceedings of Machine Learning Research},
  publisher = {PMLR},
  url = {https://proceedings.mlr.press/v235/gatmiry24b.html}
}

@misc{zhu2026scalinglatentreasoninglooped,
      title={Scaling Latent Reasoning via Looped Language Models},
      author={Rui-Jie Zhu and Zixuan Wang and Kai Hua and Tianyu Zhang and Ziniu Li and Haoran Que and Boyi Wei and Zixin Wen and Fan Yin and He Xing and Lu Li and Jiajun Shi and Kaijing Ma and Shanda Li and Taylor Kergan and Andrew Smith and Xingwei Qu and Mude Hui and Bohong Wu and Qiyang Min and Hongzhi Huang and Xun Zhou and Wei Ye and Jiaheng Liu and Jian Yang and Yunfeng Shi and Chenghua Lin and Enduo Zhao and Tianle Cai and Ge Zhang and Wenhao Huang and Yoshua Bengio and Jason Eshraghian},
      year={2026},
      eprint={2510.25741},
      archivePrefix={arXiv},
      primaryClass={cs.CL},
      url={https://arxiv.org/abs/2510.25741},
  howpublished = {arXiv preprint arXiv:2510.25741}
}

@inproceedings{bae2025relaxedrecursivetransformerseffective,
      title={Relaxed Recursive Transformers: Effective Parameter Sharing with Layer-wise {LoRA}},
      author={Sangmin Bae and Adam Fisch and Hrayr Harutyunyan and Ziwei Ji and Seungyeon Kim and Tal Schuster},
      year={2025},
      eprint={2410.20672},
      archivePrefix={arXiv},
      primaryClass={cs.CL},
      url={https://openreview.net/forum?id=WwpYSOkkCt},
  booktitle = {International Conference on Learning Representations}
}

@misc{prairie2026parcaescalinglawsstable,
      title={Parcae: Scaling Laws For Stable Looped Language Models},
      author={Hayden Prairie and Zachary Novack and Taylor Berg-Kirkpatrick and Daniel Y. Fu},
      year={2026},
      eprint={2604.12946},
      archivePrefix={arXiv},
      primaryClass={cs.LG},
      url={https://arxiv.org/abs/2604.12946},
  howpublished = {arXiv preprint arXiv:2604.12946}
}

@inproceedings{zheng2024opencodeinterpreter,
  title={{OpenCodeInterpreter}: Integrating Code Generation with Execution and Refinement},
  author={Zheng, Tianyu and Zhang, Ge and Shen, Tianhao and Liu, Xueling and Lin, Bill Yuchen and Fu, Jie and Chen, Wenhu and Yue, Xiang},
  year={2024},
  booktitle = {Findings of the Association for Computational Linguistics: ACL 2024},
  url = {https://aclanthology.org/2024.findings-acl.762/},
  doi = {10.18653/v1/2024.findings-acl.762},
  pages = {12834--12859}
}

@misc{wang2025hierarchicalreasoningmodel,
      title={Hierarchical Reasoning Model},
      author={Guan Wang and Jin Li and Yuhao Sun and Xing Chen and Changling Liu and Yue Wu and Meng Lu and Sen Song and Yasin Abbasi Yadkori},
      year={2025},
      eprint={2506.21734},
      archivePrefix={arXiv},
      primaryClass={cs.AI},
      url={https://arxiv.org/abs/2506.21734},
  howpublished = {arXiv preprint arXiv:2506.21734}
}

@misc{jolicoeurmartineau2025morerecursivereasoningtiny,
      title={Less is More: Recursive Reasoning with Tiny Networks},
      author={Alexia Jolicoeur-Martineau},
      year={2025},
      eprint={2510.04871},
      archivePrefix={arXiv},
      primaryClass={cs.LG},
      url={https://arxiv.org/abs/2510.04871},
  howpublished = {arXiv preprint arXiv:2510.04871}
}

@inproceedings{sahoo2024simpleeffectivemaskeddiffusion,
      title={Simple and Effective Masked Diffusion Language Models},
      author={Subham Sekhar Sahoo and Marianne Arriola and Yair Schiff and Aaron Gokaslan and Edgar Marroquin and Justin T Chiu and Alexander Rush and Volodymyr Kuleshov},
      year={2024},
      eprint={2406.07524},
      archivePrefix={arXiv},
      primaryClass={cs.CL},
      url = {https://arxiv.org/abs/2406.07524},
  booktitle = {Advances in Neural Information Processing Systems}
}

% =============================================================================
\clearpage
\appendix

\section{Implementation Details}
\label{app:impl}

For N-Queens and graph coloring, Last-Depth, Mean-Depth, Random-Depth, Soft-Depth, and BoD
use the same architecture and output head within each task. The supervised depth set for Mean-Depth,
Random-Depth, Soft-Depth, and BoD is $\mathcal D=\{1,\ldots,D_{\mathrm{train}}\}$.
Soft-Depth uses $\tau=0.25$ except in the temperature sweep. At each presentation of an input, a target is sampled
uniformly from its complete valid solution set, except in the withheld-solution ablation.

\begin{table}[htbp]
  \centering
  \caption{Training settings for the N-Queens and graph-coloring objective comparisons.}
  \label{tab:impl}
  \small
  \begin{tabularx}{\linewidth}{@{}l>{\raggedright\arraybackslash}X@{}}
    \toprule
    Setting & Value \\
    \midrule
    Recurrent transformation & Two shared Transformer layers per loop \\
    Width / heads / FFN width & 128 / 8 / 384 \\
    Output normalization & StableMax \\
    Training depth $\Dtr$ & 16 \\
    Optimizer & AdamAtan2 \\
    Learning rate / batch size & $10^{-4}$ / 768 \\
    Schedule & 2,000 warmup updates, then constant learning rate \\
    Weight decay / gradient clipping & 1 / norm 1 \\
    EMA decay & 0.999 \\
    Training updates & N-Queens: 20,000; graph coloring: 10,000 \\
    \bottomrule
  \end{tabularx}
\end{table}

\subsection{N-Queens Completion}
\label{app:nqueens}

\paragraph{Dataset.}
Each input specifies five fixed queens on a $10\times10$ board. The training, development, and test splits contain
24,320, 400, and 2,000 inputs, respectively, with equal numbers of single- and multi-solution
inputs in each split. All valid completions are enumerated, and $D_4$-equivalent inputs do not
cross splits. Reported coverage and success use the 1,000 multi-solution test inputs. Distinct
completed boards count as different solutions; outputs are not merged under $D_4$ symmetry.

\paragraph{Model and training.}
The model represents the board with 100 cell tokens and uses the shared recurrent architecture
in Table~\ref{tab:impl}. For the objective comparisons, we train each model for 20,000 updates and evaluate its EMA
weights at that point.
Evaluation collects the intermediate predictions of one trajectory through depth 64.

\subsection{Graph Coloring}
\label{app:coloring}

\paragraph{Dataset.}
Inputs are connected graphs with 10 vertices, at least 10 edges, and 2--128 valid 3-coloring
families. The training, development, and test splits contain 4,096, 512, and 512 graphs,
respectively, with graph-isomorphic inputs separated across splits. All canonical solution
families are enumerated; colorings related by a global permutation of color names belong to the same family. Every test
graph has multiple valid solutions. Coverage is computed using each graph's number of valid
families and then averaged over graphs. The verifier checks edge constraints and color range
independently of the sampled teacher.

\paragraph{Model and training.}
A projection of adjacency rows produces 10 vertex tokens. Only the 10 vertex colors contribute
to the loss; trailing padding positions in the serialized representation are excluded. The model
uses the shared configuration in Table~\ref{tab:impl}, trains for 10,000 updates, and is evaluated
using the EMA weights at that fixed update. The BoD objective minimizes graph-level mean
cross-entropy over depths, after averaging over the valid output positions. Evaluation collects
all greedy predictions from one trajectory through depth 64.

\paragraph{Temperature sweep.}
\label{app:temperature}
Figure~\ref{fig:temperature} varies $\tau=0.1,0.2,\ldots,1.0$ while retaining the graph-coloring
configuration above. Each evaluation collects all outputs from one 64-step trajectory.

\subsection{Code Generation}
\label{app:code}

\paragraph{Fine-tuning.}
We use the 3.565B-parameter Huginn model and Python examples from CodeFeedback
\citep{geiping2025huginn,zheng2024opencodeinterpreter}. The training and validation sets contain
20,000 and 1,000 examples, respectively. Examples whose combined prompt and target exceed
1,024 tokens are excluded. Table~\ref{tab:huginn} summarizes the training settings.

\begin{table}[htbp]
  \centering
  \caption{Shared fine-tuning settings for the five code-generation objectives.}
  \label{tab:huginn}
  \small
  \begin{tabularx}{\linewidth}{@{}l>{\raggedright\arraybackslash}X@{}}
    \toprule
    Setting & Value \\
    \midrule
    Model & Huginn 3.565B, shared recurrent parameters \\
    Data & CodeFeedback Python examples; 20,000 train / 1,000 validation \\
    Maximum sequence length & 1,024 tokens, including prompt and target \\
    Training duration & 1 epoch, 625 updates \\
    Effective batch size & 32 \\
    Maximum training depth & $\Dtr=32$ \\
    Optimizer & AdamW, $\beta=(0.9,0.95)$, $\epsilon=10^{-8}$, weight decay 0 \\
    Peak learning rate & $5\times10^{-7}$ \\
    Schedule & Warmup--Stable--Decay; first 63 updates warm up, final 63 decay \\
    \bottomrule
  \end{tabularx}
\end{table}

The loss at each depth is the mean teacher-forced cross-entropy over the entire response,
excluding prompt and padding tokens, as in Eq.~\ref{eq:sequence}. Last-Depth uses the loss at
depth 32. For \ourmethod{}, all depths from 1 to 32 are scored along one recurrent trajectory,
and a single depth with the lowest response-level loss is selected for each example. This depth
is then recomputed for backpropagation through its last at most eight recurrent updates; exact
loss ties are broken randomly. The selected depth is shared by all response tokens in the example,
not chosen separately for each token.

Mean-Depth averages response-level losses over depths 1 through 32;
Random-Depth samples one depth uniformly per example, shared by all response tokens.
Soft-Depth uses Eq.~\ref{eq:softdepth} with $\tau=0.25$ and $\mathcal D=\{1,\ldots,32\}$,
applied to the response-level losses. The loss is computed per example and then averaged over the batch.
All five objectives use the dataset and optimization settings in Table~\ref{tab:huginn},
with the final checkpoint at update 625.

\paragraph{Generation and scoring.}
We evaluated all 164 HumanEval and 500 MBPP problems. Each model generates one
program greedily at each fixed depth in $\{4,8,12,16,24,32\}$, giving six candidates per problem.
There is no stochastic sampling, test-guided repair, or regeneration. Prompt and generated output
together are limited to 1,024 tokens.

\section{Graph-Coloring Trajectory Analysis}
\label{app:analysis-protocol}
Figure~\ref{fig:analysis} uses the Last-Depth and BoD checkpoints trained at depth 16 with uniformly
sampled targets for 10,000 updates. We evaluate them greedily through depth 64 without input
permutations or initial-state noise.
Solution IDs denote canonical coloring families, with colorings related by global color-label
permutations assigned the same ID.
The new-solution indicator at depth $d$ is
\begin{equation}
  \1\big[\hat y_d(x)\in\mathcal Y^\star(x)\;\text{and}\;
          \hat y_d(x)\notin S_{d-1}(x)\big],
\end{equation}
where $S_0(x)=\varnothing$ and membership is evaluated using canonical color families.
Discovery indicators are checked against the preceding decoded outputs and enumerated valid
families. Each population curve averages all 512 graphs within each seed.

Relative displacement is computed before PCA as
\begin{equation}
 \Delta_d=\frac{\|\tilde h_d-\tilde h_{d-1}\|_2}{\|\tilde h_{d-1}\|_2},\qquad d=2,\ldots,64,
\end{equation}
where $\tilde h_d$ concatenates the ten recurrent core states at depth $d$. To construct the histograms of loss-minimizing depths, we find the depth with the lowest
graph-level mean loss for each graph--target pair among the 16 training depths. We weight graphs
equally and valid targets uniformly within each graph. When losses tie exactly, we select the
earliest depth attaining the minimum.

\section{Additional Results}
\label{app:full}

\subsection{Solution Discovery across Depth}
Tables~\ref{tab:controlled}--\ref{tab:coloring-current-success} give the numerical results for
Figure~\ref{fig:depth}. The evaluation sets and training budgets are specified in
Appendices~\ref{app:nqueens} and~\ref{app:coloring}.
All models are trained at depth 16 and read out greedily at every depth up to the stated budget.

\begin{table}[!htbp]
\centering
\caption{N-Queens Coverage@Depth (\%) on 1,000 multi-solution test problems. Models are trained at $D_{\mathrm{train}}=16$.}
\label{tab:controlled}
\small
\setlength{\tabcolsep}{4pt}
\begin{tabular}{lcccccc}
\toprule
Method & @1 & @4 & @8 & @16 & @32 & @64 \\
\midrule
Last-Depth & 0.00\std{0.00} & 0.00\std{0.00} & 0.00\std{0.00} & 22.27\std{1.27} & 43.29\std{3.12} & 45.75\std{3.93} \\
Mean-Depth & 0.00\std{0.00} & 2.51\std{0.18} & 20.42\std{1.20} & 36.52\std{2.21} & 44.52\std{1.66} & 49.94\std{1.05} \\
Random-Depth & 0.00\std{0.00} & 2.46\std{0.11} & 19.93\std{0.53} & 35.42\std{0.91} & 44.15\std{1.24} & 49.63\std{1.34} \\
Soft-Depth & 0.00\std{0.00} & 2.63\std{0.37} & 21.53\std{0.82} & 38.99\std{0.79} & 48.26\std{0.83} & 53.01\std{1.47} \\
Best-of-Depth & 0.00\std{0.00} & 0.10\std{0.09} & 17.11\std{1.25} & 85.92\std{0.93} & 90.97\std{0.98} & 91.03\std{1.05} \\
\bottomrule
\end{tabular}
\end{table}

\begin{table}[!htbp]
\centering
\caption{N-Queens Success@Depth (\%) on 1,000 multi-solution test problems. Models are trained at $D_{\mathrm{train}}=16$.}
\label{tab:nq-success}
\small
\setlength{\tabcolsep}{4pt}
\begin{tabular}{lcccccc}
\toprule
Method & @1 & @4 & @8 & @16 & @32 & @64 \\
\midrule
Last-Depth & 0.00\std{0.00} & 0.00\std{0.00} & 0.00\std{0.00} & 44.80\std{2.52} & 78.23\std{4.48} & 81.13\std{4.94} \\
Mean-Depth & 0.00\std{0.00} & 5.10\std{0.40} & 40.37\std{1.83} & 65.97\std{3.25} & 75.37\std{2.50} & 81.97\std{2.46} \\
Random-Depth & 0.00\std{0.00} & 5.00\std{0.20} & 39.43\std{1.34} & 64.50\std{1.61} & 75.27\std{1.72} & 82.30\std{2.35} \\
Soft-Depth & 0.00\std{0.00} & 5.30\std{0.70} & 42.37\std{1.11} & 68.67\std{1.69} & 79.13\std{2.10} & 84.20\std{2.49} \\
Best-of-Depth & 0.00\std{0.00} & 0.20\std{0.17} & 33.03\std{2.20} & 97.13\std{0.55} & 99.00\std{0.10} & 99.03\std{0.12} \\
\bottomrule
\end{tabular}
\end{table}

\begin{table}[!htbp]
\centering
\caption{Graph-coloring Coverage@Depth (\%) on 512 test graphs. Models are trained at $D_{\mathrm{train}}=16$.}
\label{tab:coloring-current}
\small
\setlength{\tabcolsep}{4pt}
\begin{tabular}{lcccccc}
\toprule
Method & @1 & @4 & @8 & @16 & @32 & @64 \\
\midrule
Last-Depth & 0.05\std{0.01} & 2.01\std{0.10} & 8.55\std{0.77} & 15.79\std{0.88} & 18.30\std{0.88} & 18.62\std{0.76} \\
Mean-Depth & 2.67\std{0.09} & 11.99\std{0.25} & 14.11\std{0.05} & 14.84\std{0.03} & 15.09\std{0.07} & 15.15\std{0.07} \\
Random-Depth & 2.38\std{0.20} & 11.51\std{0.15} & 13.87\std{0.29} & 14.42\std{0.25} & 14.68\std{0.15} & 14.75\std{0.16} \\
Soft-Depth & 0.88\std{0.25} & 15.17\std{0.59} & 24.36\std{0.66} & 36.64\std{0.53} & 42.18\std{0.77} & 44.39\std{0.80} \\
Best-of-Depth & 0.71\std{0.42} & 14.18\std{0.53} & 30.76\std{0.70} & 54.95\std{0.80} & 66.79\std{1.89} & 72.16\std{3.62} \\
\bottomrule
\end{tabular}
\end{table}

\begin{table}[!htbp]
\centering
\caption{Graph-coloring Success@Depth (\%) on 512 test graphs. Models are trained at $D_{\mathrm{train}}=16$.}
\label{tab:coloring-current-success}
\small
\setlength{\tabcolsep}{4pt}
\begin{tabular}{lcccccc}
\toprule
Method & @1 & @4 & @8 & @16 & @32 & @64 \\
\midrule
Last-Depth & 0.59\std{0.34} & 20.70\std{1.47} & 50.98\std{2.92} & 70.31\std{1.86} & 73.76\std{1.39} & 74.09\std{1.30} \\
Mean-Depth & 23.37\std{0.96} & 61.72\std{1.22} & 67.38\std{0.20} & 68.42\std{0.11} & 68.82\std{0.11} & 68.88\std{0.11} \\
Random-Depth & 21.88\std{1.28} & 60.16\std{0.52} & 66.28\std{0.63} & 67.12\std{0.30} & 67.32\std{0.11} & 67.38\std{0.00} \\
Soft-Depth & 10.74\std{1.79} & 79.88\std{2.34} & 94.60\std{1.58} & 97.98\std{0.11} & 98.44\std{0.20} & 98.63\std{0.20} \\
Best-of-Depth & 9.18\std{4.04} & 74.48\std{1.47} & 94.21\std{0.74} & 99.67\std{0.30} & 99.87\std{0.23} & 99.93\std{0.11} \\
\bottomrule
\end{tabular}
\end{table}

At depth 64, graph-coloring BoD achieves $99.93\pm0.11\%$ success and
$72.16\pm3.62\%$ coverage. On the multi-solution N-Queens test population, the corresponding
values are $99.03\pm0.12\%$ and $91.03\pm1.05\%$.

\subsection{Width and Depth under a Fixed Recurrent-Step Budget}
Tables~\ref{tab:nq-budget} and~\ref{tab:gc-budget} give the values plotted in Figure~\ref{fig:compute-depth}(a, b).
All seven allocations use $ND=64$, pooling valid predictions from every intermediate depth of
all trajectories before computing per-problem coverage. Each training seed averages 32 inference repeats.

For each N-Queens trajectory, we initialize the recurrent state with independent Gaussian noise
scaled by the encoded input:
\begin{equation}
 h_0^{(j)}=0.1\,\mathrm{RMS}(\Enc_\theta(x))\,\epsilon_j,
 \qquad \epsilon_j\sim\mathcal N(0,I),
\end{equation}
where RMS is the root mean square over entries of the encoded input, which remains unchanged
during initialization. The scale is selected from $\{0,0.03,0.1,0.3\}$ by mean development
coverage at $(N,D)=(4,16)$ across both methods and all three seeds. The selected value, 0.1,
is fixed before test evaluation. For graph coloring, we instead independently permute vertices for each trajectory, map predictions
back to the original vertex order, and canonicalize color names before pooling solutions.
Both procedures also apply at $N=1$.

\begin{table}[!htbp]
\centering
\caption{N-Queens Coverage@Depth (\%) at $ND=64$, pooling all $N$ trajectories through depth $D$.}
\label{tab:nq-budget}
\small
\setlength{\tabcolsep}{4pt}
\begin{tabular}{lccc}
\toprule
Depth $D$ & Trajectories $N$ & Last-Depth & Best-of-Depth \\
\midrule
1 & 64 & 0.00\std{0.00} & 0.00\std{0.00} \\
2 & 32 & 0.00\std{0.00} & 0.00\std{0.00} \\
4 & 16 & 0.00\std{0.00} & 0.50\std{0.30} \\
8 & 8 & 0.00\std{0.00} & 31.78\std{1.81} \\
16 & 4 & 48.49\std{2.06} & 94.13\std{0.56} \\
32 & 2 & 56.25\std{2.40} & 93.45\std{0.92} \\
64 & 1 & 44.55\std{3.85} & 88.28\std{1.32} \\
\bottomrule
\end{tabular}
\end{table}

\begin{table}[!htbp]
\centering
\caption{Graph-coloring Coverage@Depth (\%) at $ND=64$, pooling all $N$ trajectories through depth $D$.}
\label{tab:gc-budget}
\small
\setlength{\tabcolsep}{4pt}
\begin{tabular}{lccc}
\toprule
Depth $D$ & Trajectories $N$ & Last-Depth & Best-of-Depth \\
\midrule
1 & 64 & 2.25\std{1.25} & 33.85\std{10.34} \\
2 & 32 & 2.26\std{1.35} & 62.86\std{2.63} \\
4 & 16 & 20.61\std{1.82} & 77.28\std{0.67} \\
8 & 8 & 36.39\std{0.65} & 82.80\std{0.20} \\
16 & 4 & 37.50\std{0.51} & 85.86\std{0.49} \\
32 & 2 & 28.54\std{0.52} & 82.67\std{1.44} \\
64 & 1 & 18.12\std{0.41} & 72.14\std{3.47} \\
\bottomrule
\end{tabular}
\end{table}

\subsection{Training Depth and Test-Time Extrapolation}
The training-depth ablation uses BoD models trained at depths 4, 8, and 16, each with three seeds.
The protocol holds the evaluation population and training-update budget fixed within each task,
as specified in Appendix~\ref{app:impl}. Depth-16 models are the same as in
Figure~\ref{fig:depth}. Evaluation uses unperturbed greedy trajectories. Tables~\ref{tab:nq-train-depth}
and~\ref{tab:gc-train-depth} give coverage values for Figure~\ref{fig:compute-depth}(c, d).

\begin{table}[!htbp]
\centering
\caption{N-Queens BoD Coverage@Depth (\%) across training and test depths.}
\label{tab:nq-train-depth}
\small
\setlength{\tabcolsep}{4pt}
\begin{tabular}{lccccc}
\toprule
Training depth & @4 & @8 & @16 & @32 & @64 \\
\midrule
4 & 6.48\std{3.88} & 8.11\std{3.96} & 8.21\std{4.01} & 8.21\std{4.01} & 8.21\std{4.01} \\
8 & 6.61\std{2.73} & 60.21\std{11.72} & 75.15\std{11.99} & 75.66\std{11.90} & 75.70\std{11.95} \\
16 & 0.10\std{0.09} & 17.11\std{1.25} & 85.92\std{0.93} & 90.97\std{0.98} & 91.03\std{1.05} \\
\bottomrule
\end{tabular}
\end{table}

\begin{table}[!htbp]
\centering
\caption{Graph-coloring BoD Coverage@Depth (\%) across training and test depths.}
\label{tab:gc-train-depth}
\small
\setlength{\tabcolsep}{4pt}
\begin{tabular}{lccccc}
\toprule
Training depth & @4 & @8 & @16 & @32 & @64 \\
\midrule
4 & 21.09\std{0.50} & 27.64\std{0.83} & 31.53\std{0.54} & 33.32\std{0.64} & 34.13\std{0.69} \\
8 & 17.02\std{0.55} & 34.69\std{0.92} & 44.86\std{0.90} & 50.38\std{1.10} & 53.11\std{1.57} \\
16 & 14.18\std{0.53} & 30.76\std{0.70} & 54.95\std{0.80} & 66.79\std{1.89} & 72.16\std{3.62} \\
\bottomrule
\end{tabular}
\end{table}

\begin{figure}[!htbp]
  \centering
  \includegraphics[width=\linewidth]{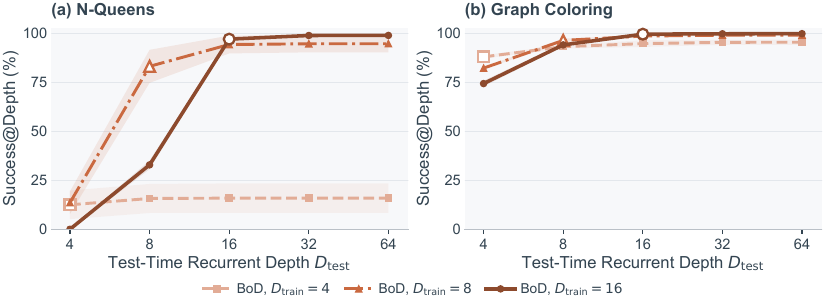}
  \caption{\textbf{Success across training and evaluation depths.} Success@Depth for BoD trained
    at depths 4, 8, and 16, using the same checkpoints and evaluation problems as Figure~\ref{fig:compute-depth}(c, d).
    Open markers indicate evaluation at the training depth; bands indicate $\pm1$ SD.}
  \label{fig:train-success}
\end{figure}

Figure~\ref{fig:train-success} and Table~\ref{tab:train-depth-success} show the corresponding success
results. Success increases beyond the training depth for the depth-4 and depth-8 models on both tasks.
For graph coloring, success saturates earlier than coverage, separating initial solution finding
from continued discovery of alternatives.

\begin{table}[!htbp]
\centering
\caption{BoD Success@Depth (\%) for Figure~\ref{fig:train-success}.}
\label{tab:train-depth-success}
\small
\setlength{\tabcolsep}{4pt}
\begin{tabular}{lccccc}
\toprule
Training depth & @4 & @8 & @16 & @32 & @64 \\
\midrule
\multicolumn{6}{c}{N-Queens} \\
4 & 12.67\std{7.31} & 15.90\std{7.41} & 16.07\std{7.51} & 16.07\std{7.51} & 16.07\std{7.51} \\
8 & 13.73\std{5.78} & 83.27\std{8.42} & 94.40\std{4.54} & 94.77\std{4.37} & 94.80\std{4.39} \\
16 & 0.20\std{0.17} & 33.03\std{2.20} & 97.13\std{0.55} & 99.00\std{0.10} & 99.03\std{0.12} \\
\multicolumn{6}{c}{Graph Coloring} \\
4 & 88.09\std{0.85} & 93.29\std{0.92} & 94.86\std{1.71} & 95.51\std{1.28} & 95.57\std{1.30} \\
8 & 82.36\std{0.56} & 96.48\std{0.90} & 98.70\std{0.45} & 99.15\std{0.30} & 99.22\std{0.39} \\
16 & 74.48\std{1.47} & 94.21\std{0.74} & 99.67\std{0.30} & 99.87\std{0.23} & 99.93\std{0.11} \\
\bottomrule
\end{tabular}
\end{table}

\subsection{Sensitivity to the Soft-Depth Temperature}
\label{app:temperature-results}

\begin{figure}[!htbp]
  \centering
  \includegraphics[width=\linewidth]{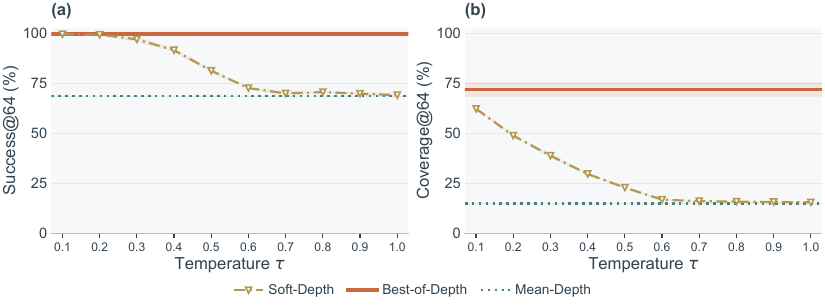}
  \caption{\textbf{Sensitivity to the Soft-Depth temperature on graph coloring.}
    Test success and coverage use all greedy readouts through depth 64. The curve varies
    $\tau=0.1,0.2,\ldots,1.0$; horizontal lines show BoD and Mean-Depth under the same configuration.
    Bands indicate $\pm1$ SD.}
  \label{fig:temperature}
\end{figure}

The temperature in Soft-Depth controls how strongly learning concentrates on the lowest-loss
readouts. We vary $\tau\in\{0.1,0.2,\ldots,1.0\}$ on graph coloring and evaluate cumulative success
and coverage through depth 64, keeping all other settings fixed (Appendix~\ref{app:temperature}).
Coverage decreases as temperature increases, approaching
Mean-Depth at larger temperatures (Figure~\ref{fig:temperature}). The development-selected
$\tau=0.1$ achieves 62.41\% test coverage, compared with 72.16\% for BoD, while both achieve
success above 99\%.

\subsection{Withheld-Solution Discovery Protocol}
\label{app:teacher-diversity}

The ablation in Section~\ref{sec:unseen-solutions} uses 3,034 graphs from the graph-coloring
training split with at least five canonical valid solutions per graph. The cohort contains
79,713 solutions in total, with 5--128 per graph (mean 26.27). This restriction leaves
withheld solutions in every one- through four-target condition; it changes the evaluation cohort
relative to the full graph-coloring experiment. All coloring families are identified up to a
global permutation of color names, as in Appendix~\ref{app:coloring}.

For each graph and experimental seed, one random ordering of all valid solutions defines nested
teacher pools of sizes one, two, three, and four. The all-solutions condition uses the complete
pool. Pools remain fixed throughout each run; only the selected teacher is resampled at each
presentation. Last-Depth and BoD share teacher pools, target draws, graph order, and initialization
within each seed. The protocol uses the architecture and optimizer in Appendix~\ref{app:coloring},
training depth 16, 10,000 updates, and three paired seeds.

Last-Depth and BoD each run one unperturbed trajectory through depth 64. Last-Depth + BoN reuses
the Last-Depth checkpoint and pools predictions from independently vertex-permuted trajectories
under $ND=64$. The allocation is selected from the seven pairs in Section~\ref{sec:widthdepth}
using development-set coverage averaged across teacher conditions and seeds. This selects
$N=4$ trajectories of depth $D=16$, fixed before evaluating withheld-solution discovery. Outputs are mapped back to the original vertex order
before verification and canonicalization. Every method collects all intermediate readouts;
Last-Depth + BoN therefore denotes sampling with all-depth readout, rather than terminal-only
sampling. Randomized evaluation uses 32 repeats per training seed.

Seen and unseen coverage are macro-averaged over the same fixed graph cohort, using the teacher
pool and its valid-solution complement as their respective denominators. Since increasing
the teacher pool reduces the number of withheld solutions, supporting measurements include total coverage, the
number of distinct unseen solutions recovered and the fraction of graphs with at least one such
solution. The all-solutions condition has no withheld solutions, so its unseen coverage is
undefined, while seen coverage equals total coverage.

\subsection{Additional Baseline Protocols}
\label{app:other-baselines}

The comparisons in Table~\ref{tab:other-baselines} use the same datasets, teacher sampling, optimization settings,
width 128, eight attention heads, and FFN width 384 as the main experiments. Each method
uses three training seeds and a fixed final EMA checkpoint after 20,000 updates for N-Queens
or 10,000 updates for graph coloring. These are independent implementations under a common
training recipe. Parameter counts and training and inference FLOPs are not matched.

\begin{table}[!htbp]
\centering
\caption{Baseline comparison on N-Queens and graph coloring. Success and Coverage (\%) summarize each method's candidate set. Ouro uses an instance-level adaptation of its training objective. Compute and candidate counts differ across architectures; Appendix~\ref{app:other-baselines} gives the protocol.}
\label{tab:other-baselines}
\small
\setlength{\tabcolsep}{4pt}
\renewcommand{\arraystretch}{1.12}
\begin{tabular*}{\linewidth}{@{\extracolsep{\fill}}lcccc@{}}
\toprule
& \multicolumn{2}{c}{N-Queens} & \multicolumn{2}{c}{Graph Coloring} \\
\cmidrule(lr){2-3}\cmidrule(l){4-5}
Method & Success & Coverage & Success & Coverage \\
\midrule

Transformer (8 layers) & 9.8{\scriptsize$\,\pm\,$1.3} & 4.8{\scriptsize$\,\pm\,$0.6} & 39.6{\scriptsize$\,\pm\,$1.8} & 5.2{\scriptsize$\,\pm\,$0.5} \\
Transformer (32 layers) & 10.9{\scriptsize$\,\pm\,$1.1} & 5.4{\scriptsize$\,\pm\,$0.6} & 44.5{\scriptsize$\,\pm\,$1.1} & 6.3{\scriptsize$\,\pm\,$0.3} \\
Autoregressive Transformer & 57.4{\scriptsize$\,\pm\,$3.0} & 27.4{\scriptsize$\,\pm\,$1.4} & 67.9{\scriptsize$\,\pm\,$1.5} & 8.2{\scriptsize$\,\pm\,$0.4} \\
Masked Diffusion & 34.2{\scriptsize$\,\pm\,$4.1} & 16.3{\scriptsize$\,\pm\,$2.0} & 76.1{\scriptsize$\,\pm\,$1.5} & 10.3{\scriptsize$\,\pm\,$0.3} \\
HRM & 98.6{\scriptsize$\,\pm\,$0.5} & 48.0{\scriptsize$\,\pm\,$0.6} & 67.4{\scriptsize$\,\pm\,$0.9} & 13.7{\scriptsize$\,\pm\,$0.2} \\
TRM & 49.4{\scriptsize$\,\pm\,$23.6} & 23.8{\scriptsize$\,\pm\,$11.4} & 67.4{\scriptsize$\,\pm\,$1.5} & 14.3{\scriptsize$\,\pm\,$0.3} \\
Ouro & 77.9{\scriptsize$\,\pm\,$4.5} & 46.8{\scriptsize$\,\pm\,$3.7} & 73.3{\scriptsize$\,\pm\,$1.3} & 18.6{\scriptsize$\,\pm\,$0.1} \\
\midrule
\textbf{Best-of-Depth} & \textbf{99.0}{\scriptsize$\,\pm\,$0.1} & \textbf{91.0}{\scriptsize$\,\pm\,$1.0} & \textbf{99.9}{\scriptsize$\,\pm\,$0.1} & \textbf{72.2}{\scriptsize$\,\pm\,$3.6} \\
\bottomrule
\end{tabular*}
\end{table}

The unshared Transformers have eight or 32 layers and produce one candidate per input.
The autoregressive baseline uses a four-layer decoder and produces one complete candidate.
Masked diffusion uses four layers with a SUBS parameterization and a completed 64-step
denoising schedule. HRM uses four-layer low- and high-level modules with two cycles at each
level; TRM uses two shared layers, four inner iterations, three cycles, and at most 16 outer
training segments. HRM and TRM are evaluated with a fixed 64-step expansion rather than their
native adaptive stopping rule. BoD and Ouro also collect all readouts through depth 64.
Success and coverage are computed on the resulting candidate sets, using the definitions
in Section~\ref{sec:setup}. For the autoregressive and diffusion baselines, metrics are averaged
across three inference seeds within each training seed, without pooling candidates across
repeats.

The Ouro baseline applies its Stage I expected-loss and entropy-regularization objective to
whole instances, using the same looped backbone as BoD. A shared linear gate pools the recurrent
states over unknown cells for N-Queens and over vertices for graph coloring, then predicts
a sigmoid exit hazard at each depth. The first 15 hazards define an exit distribution by
survival products, with the remaining mass assigned to depth 16. The gate adds 129 parameters;
its weights start at zero and its bias at $-\log 15$. Training minimizes the expected
per-instance loss plus $\beta$ times the KL divergence of the exit distribution from the
uniform distribution over the 16 training depths. Evaluation collects all greedy readouts
through depth 64; the gate neither stops the trajectory nor filters candidates. This comparison
adapts the learning objective, rather than evaluating a pretrained Ouro language model or
reproducing its second training stage.

We select one positive coefficient per task by mean development Coverage@Depth at depth 64
over three training seeds, using the grid $\{0.001,0.003,0.01,0.03,0.05,0.1,0.25,1\}$.
A separate $\beta=0$ control is excluded from selection; exact ties favor the smaller coefficient.
The selected values are $\beta=0.03$ for N-Queens and $\beta=0.001$ for graph coloring, using
200 multi-solution development puzzles and 512 development graphs, respectively.

\section{Theoretical Analysis}
\label{app:theory}

Throughout this appendix, fix a non-autoregressive input $x$ and a deterministic recurrent
trajectory. Targets are drawn from the finite valid solution set $\mathcal Y^\star(x)$, with
cardinality $M\ge1$, uniformly unless otherwise specified. For the cross-entropy results, all depths
$1,\ldots,D$ are read out. The joint readout distribution on the
complete output space $\Omega$, with $p_{d,j}$ the categorical distribution at predicted position $j$, is
\begin{equation}
 P_d(y\mid x)=\prod_{j=1}^{T(x)}p_{d,j}(y_j\mid x).
 \label{eq:joint-readout}
\end{equation}
It is normalized over $\Omega$, which includes invalid assignments. All logarithms are natural.
For cross-entropy, define the expected whole-example loss
\begin{equation}
 s_D(x)=T(x)\,\mathbb E_Y[\cL_{\text{BoD}}(x,Y)]
       =\mathbb E_Y\!\left[-\log\max_{d\le D}P_d(Y\mid x)\right].
 \label{eq:bod-risk}
\end{equation}
Here $T(x)$ is the number of predicted positions; the factor $T(x)$ reverses the position averaging
in the training loss. We suppress $x$ where the conditioning is unambiguous.

\subsection{Proofs of Propositions~\ref{prop:firsthit} and~\ref{prop:coverage}}
\label{app:exact-match-proof}

For Proposition~\ref{prop:firsthit}, the minimum of the binary losses is zero exactly when at
least one readout equals the target, which is equivalent to $\tau_y\le D$. Hence the loss equals
$\1[\tau_y>D]$ pointwise. Taking an expectation over any distribution of target--trajectory pairs
gives Eq.~\ref{eq:firsthit}.

For Proposition~\ref{prop:coverage}, fix the candidate set independently of the sampled target.
For any distribution $q$ supported on $\mathcal Y^\star(x)$,
\begin{align}
 \mathbb E_Y[\cL_{\mathrm{BoD}}^{0\text{--}1}\mid x]
 &=\sum_{y\in\mathcal Y^\star(x)}q(y\mid x)\,\1[y\notin S_D(x)]\\
 &=1-\sum_{y\in S_D(x)\cap\mathcal Y^\star(x)}q(y\mid x).
\end{align}
When $q(y\mid x)=1/M$, the second term is exactly the fraction of valid solutions in the
candidate set, proving Eq.~\ref{eq:coverage-risk}. For a random candidate set, the same identity
holds conditional on that set, provided it is independent of the target given $x$; averaging
then yields expected coverage.

For graph-coloring family coverage, both the candidates in $S_D(x)$ and the predictions in the
exact-match loss are canonicalized in the same way. The target set contains one canonical
representative per valid family. Using raw color-label equality on the left and family coverage
on the right would not give this identity. The cross-entropy results below instead use the raw
joint distribution in Eq.~\ref{eq:joint-readout}: generating a canonical target is sufficient to
discover its family, so their coverage lower bounds also hold after canonicalization.

\subsection{Depth-Wise Sampling and Coverage}
\label{app:discovery-proof}

\begin{lemma}[Distinct valid solutions from depth-wise sampling]
\label{lem:discovery}
Draw $\tilde Y_d\sim P_d(\cdot\mid x)$ at each depth and set $S_D(x)=\{\tilde Y_d:d\le D\}$.
Let $K_D$ count the distinct valid solutions in this random candidate set. Then
\begin{equation}
 \mathbb E[K_D\mid x]\ge M e^{-s_D(x)},\qquad
 \mathbb E[\mathrm{Coverage@Depth}(D;x)\mid x]\ge e^{-s_D(x)}.
 \label{eq:sampled-count}
\end{equation}
The expectations are over candidate sampling; independence between candidates is not required.
\end{lemma}

\begin{proof}
For each valid solution $y$, define $A_y=\{\tilde Y_d=y\text{ for some }d\le D\}$. The union
contains each individual event, hence
\begin{equation}
 \Pr(A_y\mid x)\ge\max_{d\le D}\Pr(\tilde Y_d=y\mid x)
                  =\max_{d\le D}P_d(y\mid x).
\end{equation}
This uses only the marginal distribution of each candidate. Let
\begin{equation}
 a_D(y)=\max_{d\le D}P_d(y\mid x),\qquad
 Z_D^\star=\sum_{y\in\mathcal Y^\star(x)}a_D(y).
 \label{eq:valid-envelope}
\end{equation}
Linearity of expectation gives $\mathbb E[K_D\mid x]\ge Z_D^\star$. Since targets are uniform,
Jensen's inequality for the convex function $-\log$ yields
\begin{equation}
 s_D=\frac1M\sum_{y\in\mathcal Y^\star(x)}-\log a_D(y)
 \ge -\log\!\left(\frac{Z_D^\star}{M}\right).
 \label{eq:envelope-risk}
\end{equation}
Thus $Z_D^\star\ge M e^{-s_D}$, proving the count bound. Dividing by $M$ gives the coverage
bound. If some valid target has $a_D(y)=0$, then $s_D=\infty$ and the stated lower bound is zero.
\end{proof}

The result also holds when additional candidates are drawn at any depth because adding candidates
 cannot remove the discovered solution. The sampling distributions must match those used in the loss function.
For independent draws, the exact expected count for one sample per depth is
\begin{equation}
 \mathbb E[K_D\mid x]
 =\sum_{y\in\mathcal Y^\star(x)}\left[1-\prod_{d=1}^D(1-P_d(y\mid x))\right].
\end{equation}

\paragraph{Sharpness.}
Suppose $M$ is divisible by $D$. Partition the valid solutions into $D$ disjoint subsets of
size $M/D$, and let $P_d$ be uniform on the $d$th subset. Then $s_D=\log(M/D)$, and one draw
from each depth produces exactly $D=M e^{-s_D}$ distinct valid solutions. This establishes
sharpness as a bound on normalized readout distributions. It does not require that a particular
factorized decoder can represent every such partition.

\subsection{A Shared-Recurrent Construction for Proposition~\ref{prop:separation}}
\label{app:separation-proof}

\paragraph{Task and valid solutions.}
There is one input, the labeled complete bipartite graph $K_{3,3}$. Its valid outputs are the
six perfect matchings, represented by the $3\times3$ permutation matrices in $\mathcal M$.
The output space is $\{0,1\}^{3\times3}$; a selected edge has value one. A training target
$Y$ is drawn independently and uniformly from $\mathcal M$ at each update. Every edge occurs
in two matchings, so $\mathbb E[Y]=J/3$, where $J$ is the all-ones matrix. The model does not
receive the target as input, and decoding does not enforce the matching constraints.

\paragraph{One shared recurrence.}
Let $r=(1\ 2\ 3)$ permute the left vertices and $c=(1\ 2)$ permute the right vertices.
Define the fixed linear map $\mathcal R$ by
$(\mathcal R A)_{r(i),c(j)}=A_{ij}$. This is a symmetry of the graph and preserves matchings.
It satisfies $\mathcal R^6=I$, and every matching has a full orbit:
\begin{equation}
 \{\mathcal R^dM:d=1,\ldots,6\}=\mathcal M\qquad(M\in\mathcal M).
 \label{eq:matching-orbit}
\end{equation}
Indeed, identify a matching with a permutation $\pi$ from left to right vertices.
If $\mathcal R^dM^\pi=M^\pi$, then $c^d=\pi r^d\pi^{-1}$. When $3\nmid d$, the
right side has order three while the left has order one or two. Thus $3\mid d$, and the
identity then also requires $2\mid d$. The orbit therefore has length six.

The only trainable parameter is a matrix $\theta\in\mathbb R^{3\times3}$. Let $\Pi$ denote the
permutation matrix representing $\mathcal R$ on vectorized matrices. A ten-dimensional hidden
state with fixed initialization evolves according to
\begin{equation}
 h_0=\begin{pmatrix}0_9\\1\end{pmatrix},\qquad
 h_d=S(\theta)h_{d-1},\qquad
 S(\theta)=\begin{pmatrix}
 \Pi & \Pi\operatorname{vec}(\theta)\\
 0_{1\times9}&0
 \end{pmatrix}.
 \label{eq:matching-recurrence}
\end{equation}
The first nine entries of $h_d$ give a logit matrix $H_d$; the last entry is zero after the
first recurrent step. Consequently, $H_d=\mathcal R^d\theta$. The same parameter-dependent matrix
$S(\theta)$ is applied at every depth, but the trainable pattern is injected only at the first
step. Subsequent steps apply the prescribed permutation. Thus, this construction studies
how the objective uses an available recurrent orbit, rather than learning that orbit's rule.

\paragraph{Readout and common training protocol.}
All methods use $D=6$ and the shared factorized Bernoulli readout
\begin{equation}
 P_{\theta,d}(y)=\prod_{i,j}\sigma(H_{d,ij})^{y_{ij}}
 [1-\sigma(H_{d,ij})]^{1-y_{ij}},\qquad
 L_d(\theta;Y)=-\frac19\log P_{\theta,d}(Y),
 \label{eq:matching-readout}
\end{equation}
where $\sigma$ is the sigmoid. Greedy decoding selects an edge exactly when its logit is
positive. Each method starts at $\theta^{(0)}=bJ$, with $b=-\log2$, and uses single-example
SGD with $\eta_n=9/(n+1)$. This is equivalent to step size $1/(n+1)$ on the summed loss;
the factor nine accounts for the position averaging in Eq.~\ref{eq:matching-readout}.
Random-Depth samples its depth independently of the target. BoD selects one minimizing
depth uniformly when losses tie. Soft-Depth uses any fixed $\tau>0$ in Eq.~\ref{eq:softdepth}.
All methods may share the same target sequence. For the one input in this construction,
\begin{equation}
 \mathrm{Coverage@Depth}(6;\theta)
 =\frac{\big|\{\mathbf1[\mathcal R^d\theta>0]:d=1,\ldots,6\}\cap\mathcal M\big|}{6}.
 \label{eq:matching-coverage}
\end{equation}
The indicator acts entrywise. This makes explicit the parameter dependence used in
Proposition~\ref{prop:separation}.

\begin{proof}[Proof of Proposition~\ref{prop:separation}]
Write $\Phi(\theta)=\sum_{i,j}\log(1+e^{\theta_{ij}})$ and use the Frobenius inner product.
Because $\mathcal R$ permutes entries,
\begin{equation}
 L_d(\theta;Y)=\frac19\big[\Phi(\theta)-\langle\theta,\mathcal R^{-d}Y\rangle\big],
 \qquad
 \nabla_\theta L_d(\theta;Y)=\frac19\big[\sigma(\theta)-\mathcal R^{-d}Y\big].
 \label{eq:matching-loss-gradient}
\end{equation}

\textit{Baseline minimizers and stationary objectives.}
Let $R_A$ denote the expected loss for objective $A$, including any random depth.
Equation~\ref{eq:matching-orbit} implies
$\frac16\sum_{d=1}^6\mathcal R^{-d}Y=J/3$ for every target. Hence
\begin{equation}
 R_{\mathrm{Last}}(\theta)=R_{\mathrm{Mean}}(\theta)=R_{\mathrm{Random}}(\theta)
 =\frac19\big[\Phi(\theta)-\langle\theta,J/3\rangle\big].
 \label{eq:matching-baseline-risk}
\end{equation}
This risk has positive diagonal Hessian entries
$\sigma(\theta_{ij})[1-\sigma(\theta_{ij})]/9$ and the unique minimizer $bJ$.
For Mean-Depth the loss already equals this risk for each individual target, so its gradient
is zero at initialization and it remains there.

At $bJ$, all depthwise losses tie. Soft-Depth therefore assigns weight $1/6$ to each gradient,
independently of $\tau>0$. For every target,
\begin{equation}
 \left.\nabla_\theta\cL_{\mathrm{Soft-Depth}}\right|_{bJ}
 =\frac19\left[\frac13J-\frac16\sum_{d=1}^6\mathcal R^{-d}Y\right]=0.
 \label{eq:matching-soft-stationary}
\end{equation}
It too remains exactly at initialization. Every logit is then $b<0$ at every depth, so
Mean-Depth and Soft-Depth always have zero greedy coverage.

\textit{Almost-sure convergence for Last-Depth and Random-Depth.}
Their updates have the form
$\theta^{(n+1)}=\theta^{(n)}-a_n[\sigma(\theta^{(n)})-Z_n]$, where $a_n=1/(n+1)$ and,
conditional on the training history $\mathcal F_n$, $Z_n$ is uniform over $\mathcal M$.
Set $V_n=\|\theta^{(n)}-bJ\|_F^2$ and
$A(\theta)=\langle\theta-bJ,\sigma(\theta)-J/3\rangle$.
Strict monotonicity of the sigmoid gives $A(\theta)>0$ except at $bJ$; the bounded components
of the stochastic gradient give
\begin{equation}
 \mathbb E[V_{n+1}\mid\mathcal F_n]
 \le V_n-2a_nA(\theta^{(n)})+9a_n^2.
 \label{eq:matching-drift}
\end{equation}
Thus $V_n+9\sum_{k=n}^\infty a_k^2$ is a nonnegative supermartingale. It converges almost surely,
and summing its expected drift yields
$\sum_n a_n A(\theta^{(n)})<\infty$ almost surely. Consequently $V_n$ has a finite limit.
If this limit were positive, the iterates would eventually lie in a compact annulus about
$bJ$ on which $A$ has a positive minimum. This contradicts $\sum_n a_n=\infty$.
Therefore both methods converge almost surely to $bJ$. Since $b<0$, all logits eventually
remain negative, and greedy coverage is eventually zero.

\textit{BoD breaks symmetry and retains one latent matching.}
The orbit identity gives, for every target,
\begin{equation}
 \cL_{\mathrm{BoD}}(\theta;Y)
 =\frac19\left[\Phi(\theta)-\max_{M\in\mathcal M}\langle\theta,M\rangle\right].
 \label{eq:matching-bod-risk}
\end{equation}
At initialization, let the selected tied depth for $Y_0$ be $d_0^\star$ and set
$M_\star=\mathcal R^{-d_0^\star}Y_0$. The first update gives
\begin{equation}
 \theta^{(1)}=bJ+M_\star-J/3.
 \label{eq:matching-first-update}
\end{equation}
Its entries equal $\alpha_1=b+2/3$ on $M_\star$ and $\beta_1=b-1/3$ elsewhere.
For any other matching sharing $t<3$ edges with $M_\star$, the score gap is
$(3-t)(\alpha_1-\beta_1)>0$. Thus $M_\star$ is now the unique maximum-score matching.
For every subsequent target $Y$, BoD selects the phase satisfying $\mathcal R^{d^\star}M_\star=Y$, so
\begin{equation}
 \nabla_\theta\cL_{\mathrm{BoD}}(\theta;Y)=\frac19[\sigma(\theta)-M_\star].
 \label{eq:matching-alignment}
\end{equation}
All targets reinforce the same latent matching. The two logit values satisfy
\begin{equation}
 \alpha_{n+1}=\alpha_n+a_n[1-\sigma(\alpha_n)],\qquad
 \beta_{n+1}=\beta_n-a_n\sigma(\beta_n),\qquad n\ge1.
 \label{eq:matching-bod-dynamics}
\end{equation}
Their gap grows, preserving the unique maximizing matching. Both are negative after the
first update, but
$\alpha_2> -\log2+2/3+1/4>0$ while $\beta_2<0$. These signs persist.
For every $n\ge2$, greedy decoding therefore yields
$\hat y_d=\mathcal R^dM_\star$, whose six distinct values are all valid matchings.
BoD has full coverage from the second update onward. This finite-update count uses the stated
loss normalization and step sizes. Together with the baseline results, it proves the proposition.
\end{proof}

The Soft-Depth result relies on exact symmetry at initialization. The prescribed recurrent orbit
makes this an existence result for a shared recurrent model.

\paragraph{Numerical illustration.}
We run the specified SGD updates for 4,096 iterations in float64 on 64 independent teacher
streams (seeds 0--63), shared across objectives within each run. Independent random streams
select Random-Depth readouts and resolve the initial BoD tie. Soft-Depth is evaluated at
$\tau\in\{0.1,0.25,1\}$. Every reported checkpoint is decoded at all six depths and verified
against the complete set of matchings. The Soft-Depth updates are computed from their actual
weighted gradients; no special rule holds the parameters at initialization.
Table~\ref{tab:matching-separation} illustrates the proved distinction: BoD reaches full coverage
after two updates in every run, whereas all four comparison objectives have zero coverage at
the final checkpoint. Last-Depth and Random-Depth can temporarily produce valid solutions,
consistent with the theorem, which describes eventual behavior rather than behavior at every iteration.
The experiment is a numerical illustration of this construction, separate from the learned
looped Transformer experiments. The implementation and per-run results accompany the paper.

\begin{table}[!htbp]
\centering
\caption{Numerical illustration of Proposition~\ref{prop:separation}. Greedy Coverage@Depth at $D=6$ (\%), averaged over 64 independent SGD runs on the same matching problem. Columns count training updates. All objectives use the same teacher stream within each run and the same initialization and step sizes.}
\label{tab:matching-separation}
\small
\setlength{\tabcolsep}{9pt}
\begin{tabular}{lrrrrr}
\toprule
Training objective & 0 & 2 & 16 & 256 & 4,096 \\
\midrule
Last-Depth & 0.0 & 15.6 & 1.6 & 0.0 & 0.0 \\
Mean-Depth & 0.0 & 0.0 & 0.0 & 0.0 & 0.0 \\
Random-Depth & 0.0 & 18.8 & 1.6 & 0.0 & 0.0 \\
Soft-Depth ($\tau=0.1$) & 0.0 & 0.0 & 0.0 & 0.0 & 0.0 \\
Soft-Depth ($\tau=0.25$) & 0.0 & 0.0 & 0.0 & 0.0 & 0.0 \\
Soft-Depth ($\tau=1.0$) & 0.0 & 0.0 & 0.0 & 0.0 & 0.0 \\
\midrule
\textbf{Best-of-Depth} & 0.0 & 100.0 & 100.0 & 100.0 & 100.0 \\
\bottomrule
\end{tabular}
\end{table}

\subsection{A Certificate for Greedy Solution Coverage}
\label{app:greedy-proof}

Let $S_D$ contain the distinct positionwise-greedy predictions from the same readouts. Then
\begin{equation}
 \frac{|S_D\cap\mathcal Y^\star(x)|}{M}
 \ge\left[1-\frac{s_D(x)}{\log 2}\right]_+.
 \label{eq:greedy-certificate}
\end{equation}
To prove this, consider a valid target $y\notin S_D$. At every depth, at least one position is
decoded to a token other than $y_j$. At that position, $p_{d,j}(y_j\mid x)\le1/2$, so
$-\log P_d(y\mid x)\ge\log 2$. Consequently,
\begin{equation}
 \1[y\notin S_D]\le\frac{\min_{d\le D}-\log P_d(y\mid x)}{\log 2}.
\end{equation}
Averaging over the uniform target distribution proves Eq.~\ref{eq:greedy-certificate}. The result
is nontrivial when the whole-example risk is below $\log 2$; for the position-averaged loss, the
corresponding threshold is $\log 2/T(x)$. In graph coloring, matching a canonical target exactly
is sufficient to discover its canonical family, so subsequent canonicalization preserves this
lower bound.

\end{document}